\documentclass{article}
\usepackage{arxiv}

\usepackage{graphicx}%
\usepackage{multirow}%
\usepackage{amsmath,amssymb,amsfonts}%
\usepackage{amsthm}%
\usepackage{mathrsfs}%
\usepackage[title]{appendix}%
\usepackage{xcolor}%
\usepackage{textcomp}%
\usepackage{manyfoot}%
\usepackage{booktabs}%
\usepackage{algorithm}%
\usepackage{algorithmicx}%
\usepackage{algpseudocode}%
\usepackage{listings}%
\usepackage{caption}
\usepackage{subcaption}
\usepackage[utf8]{inputenc} 
\usepackage[T1]{fontenc}    
\usepackage{amsfonts}

\usepackage{url}
\usepackage{natbib}
\usepackage{mathtools}
\usepackage{makecell}
\usepackage{siunitx}
\usepackage[export]{adjustbox}

\newcommand{\be}{\begin{equation}}
\newcommand{\ee}{\end{equation}}
\newcommand{\bea}{\begin{eqnarray}}
\newcommand{\eea}{\end{eqnarray}}
\newcommand{\bne}{\begin{equation*}}
\newcommand{\ene}{\end{equation*}}
\newcommand{\bi}{\begin{itemize}}
\newcommand{\ei}{\end{itemize}}

\newcommand{\bbm}{\begin{bmatrix}}
\newcommand{\ebm}{\end{bmatrix}}
\newcommand{\bs}{\boldsymbol}
\newcommand{\bb}{\mathbb}
\newcommand{\bof}{\mathbf}
\newcommand{\mca}{\mathcal}

\newcommand{\norm}[1]{\left\lVert#1\right\rVert}

\title{A Simultaneous Approach for Training Neural Differential-Algebraic Systems of Equations}

\date{}

\renewcommand{\shorttitle}{A Simultaneous Approach for Neural DAEs}

\author{Laurens R. Lueg \thanks{Department of Chemical Engineering, Carnegie Mellon University, Pittsburgh, PA 15213.} \\
	\texttt{llueg@andrew.cmu.edu} \\
	\And
	Victor Alves $^*$\\
	\texttt{vcunhaal@andrew.cmu.edu} \\
	\And
    Daniel Schicksnus $^{*,}$\thanks{RWTH Aachen University, Process Systems Engineering (AVT.SVT), Aachen 52074, Germany} \\
    \AND
    John R. Kitchin $^*$ \\
    \texttt{jkitchin@andrew.cmu.edu} \\
    \And
	Carl D. Laird $^*$ \\
	\texttt{claird@andrew.cmu.edu} \\
    \And
	Lorenz T. Biegler $^*$ \\
	\texttt{lb01@andrew.cmu.edu} \\
}

\begin{document}
\maketitle

\begin{abstract}
Neural differential-algebraic systems of equations (DAEs) are a modeling paradigm where some unknown relationships within a DAE are modeled with a neural network and learned from data. 
Training neural DAEs is more challenging than training neural ordinary-differential equations (ODEs), particularly for higher-index systems. Existing approaches utilize differentiable pipelines, usually comprising integration, projection, operator splitting, or penalty terms for algebraic constraints. The parameters are then updated in a sequential manner using gradient descent. 
In this work, we employ the simultaneous approach for DAE-constrained parameter estimation instead. This defines a fully discretized nonlinear programming program (NLP), whose solution simultaneously obtains the neural network parameters and the trajectories of the corresponding DAE, while enforcing constraint satisfaction at the discretization points. We show that with careful initialization and handling of the neural network terms, this approach can be efficient for smaller-scale problems, including higher-index DAEs. As the number of parameters or the amount of data increase, decomposition strategies are necessary to make this method scalable. We present an approach which is inspired by existing sequential approaches. However, a tractable, discretized NLP is solved at every iteration of gradient descent and the sensitivity of its solution with respect to the neural network parameters is evaluated in an efficient manner. We demonstrate the scalability of this decomposition with respect to the number of parameters and the size of the training data set.
\end{abstract}

\section{Introduction}

The development of modeling paradigms that combine mechanistic and data-driven components has become an important task for researchers and practitioners in the area of science and engineering, promising to provide scalable computational solutions which can make use of data and incorporate first-principles-based system knowledge. 
Notable examples include physics-informed neural networks (PINNs) \citep{raissi2019}, neural ordinary differential equations (neural ODEs) \citep{chen2018neural, kidger2022neural} and universal differential equations (UDEs) \citep{rackauckas2020universal}, which have been applied in various fields, such as bioprocesses \citep{Bangi2022}, crystallization \citep{lima2025crystallization}, model-predictive control \citep{luo2023neuralmpc, casas2025comparison}, wastewater treatment \citep{huang2025waste_water}, battery modeling \citep{huang2024minn}, power systems \citep{xiao2022feasibility} and parameter estimation in process systems engineering applications \citep{Bradley2021estimation}.

A recent extension to this field of research are so-called neural differential-algebraic systems of equations (neural DAEs), which combine neural ODEs or UDEs with algebraic constraints. Depending on the application, these constraints might be known or learned from data themselves. Related topics include the reconciliation of learned models with algebraic constraints \citep{mukherjee2025development}, as well as the integration of learned system dynamics with optimal control \citep{di2024learning}. Several works discuss the use of PINNs for DAEs \citep{moya2023dae, chen2023physical, luo2025dae}. There, the numerical solution of a (known) DAE is replaced by the evaluation of one or multiple deep neural networks, which are trained using a physics-informed loss function. This differs slightly from the training of neural DAEs, the topic of this work, which aims to approximate unknown components of a DAE using neural networks.

Training algorithms for neural DAEs usually rely on computing derivatives of a loss function with respect to the parameters of the neural network through a pipeline of computational operations, including integration to deal with the differential part of the model, as well as projection  \citep{white2024projected, pal2025semi}, operator splitting \citep{koch2025learning} and/or the use of penalty terms \citep{tuor2020constrained, neary2024neural, huang2024minn, xiao2022feasibility} to handle algebraic constraints. The parameters are then updated using (stochastic) gradient descent, and the pipeline is evaluated anew; we will refer to this type of approach as sequential. Sequential approaches are often computationally scalable with respect to the amount of data, as the pipeline can be evaluated in parallel for different batches, and with respect to the number of neural network parameters, in part due to the capabilities of modern automatic differentiation (AD) implementations for deep learning. Nevertheless, gradient descent often requires manual tweaking of its parameters and many iterations to converge to a satisfactory solution -- for neural ODEs/DAEs each iteration can be costly \citep{roesch2021collocation}. 

Similar observations have been made in the past about parameter estimation problems constrained by purely mechanistic DAEs, giving rise to simultaneous solution approaches: the DAE is fully discretized, using e.g. orthogonal collocation, and a nonlinear programming solver is used to solve the resulting optimization problem \citep{biegler2007simultaneous}. This avoids the repeated call to a DAE/ODE routines, and makes use of powerful nonlinear programming solvers, such as \textsc{IPOPT} \citep{wachter2006}. DAE-constrained optimization problems can be easily modeled in continuous time with software tools such as \textsc{APMonitor} \citep{apmonitor2014},  \textsc{InfinteOpt.jl} \citep{pulsipher2022unifying} or \textsc{Pyomo.DAE} \citep{pyomoDAE}, which automatically discretize the problem and send it to a solver. Hence, in this work we investigate the use of the simultaneous approach for training neural DAEs. A similar approach has recently demonstrated promising results on small-scale neural ODEs \citep{shapovalova2025training}. For the case of DAEs, the simultaneous approach has the added advantage of being able to handle higher-index DAEs (given the use of an appropriate discretization scheme), as well as rigorously enforcing algebraic constraints, which is not guaranteed with penalty-based training methods. 

The resulting nonlinear optimization problems are potentially large, depending on the size of the neural network and the number of trajectories where data are available. Furthermore, neural network expressions are dense with respect to their trainable parameters, which increases the computational cost of solving the associated optimization problems. We demonstrate that these issues can be dealt with to successfully apply the simultaneous approach to medium-scale neural DAEs. In addition, we will show that the simultaneous approach can also be leveraged to define sequential training approaches, which retain both favorable scaling properties and aforementioned advantages related to constraint satisfaction and higher-index systems.

\subsection{Contributions}\label{sec:intro}

This work makes the following novel contributions to the field of computational optimization applied to the training neural DAEs:
\begin{itemize}
    \item The neural DAE training problem is formulated generally and the simultaneous approach is applied. This allows for consideration of a wider range of systems than previously, including higher-index DAEs and/or systems with inequality constraints.
    \item A tailored solution approach for the resulting NLP is proposed, including a specialized initialization scheme for neural DAEs, the use of Hessian approximations and the evaluation of neural network expressions using external gray-box models.
    \item A decomposition scheme which improves the scalability of the proposed approach by leveraging efficient evaluation of NLP sensitivities with respect to neural network parameters.
\end{itemize}

\subsection{Outline}
In Section \ref{sec:problem_setting}, we provide a general framework for neural DAEs and formulate the training problem which our work addresses. We then describe how the simultaneous approach is applied to this problem in Section \ref{sec:proposed_approach}, and present specific steps to make the resulting NLP more tractable. Furthermore, we discuss decomposition approaches in Section \ref{sec:decomp} and introduce a sequential method which leverages the simultaneous approach and NLP sensitivities to improve scalability. We evaluate the proposed methods on several case studies in Section \ref{sec:case_studies}, including scalability analysis. We conclude with a discussion of the strengths and limitations of our proposed methods, and related recommendations for future work in Sec. \ref{sec:conclusion}.

\section{Problem statement}\label{sec:problem_setting}

In this work, we consider the semi-explicit neural DAE defined as
\begin{subequations}\label{eq:dae_system}
\begin{align}
   \frac{d\bof{x}}{dt} &= \bof{f}(\bof{x}(t), \bof{y}(t), \bof{z}(t), \bof{p}), &&\forall_{t \in [t_0, t_f]}  \label{eq:dae_ode}\\
   0 &= \bof{h}(\bof{x}(t), \bof{y}(t), \bof{z}(t), \bof{p}), &&\forall_{t \in [t_0, t_f]} \\
   0 &= \bof{z}(t) - \bof{f}_{\text{NN}}(\bof{v}(t), \bs{\theta}), \quad \bof{v}(t) \subseteq \{\bof{x}(t), \bof{y}(t), \bof{p} \} &&\forall_{t \in [t_0, t_f]}  \label{eq:dae_nn_def}\\
   \bof{x}(t_0) &= \bof{x}_0(\bof{p}), \label{eq:dae_x0_p}
\end{align}
\end{subequations}
with differential variables $\bof{x}(t) \in \mathbb{R}^{n_x}$, algebraic variables $\bof{y}(t) \in \mathbb{R}^{n_y}$ and $\bof{z}(t) \in \mathbb{R}^{n_z}$, and independent static variables $\bof{p}\in\mathbb{R}^{n_p}$. The neural network $\bof{f}_{NN}: \mathbb{R}^{n_v + n_{\theta}} \mapsto \mathbb{R}^{n_z}$ is parametrized by $\bs{\theta} \in \mathbb{R}^{n_\theta}$ and links the unknown terms in the model, $\bof{z}(t)$, to a subset of the remaining variables, i. e., the inputs of the neural network, labeled $\bof{v}(t) \in \mathbb{R}^{n_v}$. Often, domain knowledge allows one to define a structural prior on which variables should be considered as input to the neural network.
We assume that $\bof{f} : \bb{R}^{n_x + n_y + n_z + n_p} \mapsto \bb{R}^{n_x}$, $\bof{h}: \bb{R}^{n_x + n_y + n_z + n_p} \mapsto \bb{R}^{n_y}$, and $\bof{f}_{NN}$ are Lipschitz continuous for $t \in [t_0, t_f]$, when $\bof{p}$ and $\bs{\theta}$ are specified. 
The initial state of the differential variables may depend on some of the independent static variables \eqref{eq:dae_x0_p}; indeed, the initial state might be unknown and thus included in $\bof{p}$.
To determine the index of \eqref{eq:dae_system}, \eqref{eq:dae_nn_def} is substituted into the other equations to obtain the algebraic constraint
\begin{equation}\label{eq:substituted_alg}
    0 = \bof{h}(\bof{x}(t), \bof{y}(t), \bof{f}_{\text{NN}}(\bof{x}(t), \bs{\theta}), \bof{p}).
\end{equation}
 For a given $\bof{p}$ and $\bs{\theta}$, if $\nabla_{y}\bof{h}$ is non-singular (for all $t \in [t_0, t_f]$), the DAE \eqref{eq:dae_system} is index-1. Otherwise, the index can be defined by the minimum number of differentiations of the DAE system that are necessary to obtain ODEs for the algebraic variables $\bof{y}(t)$. Thus, the analysis of the index of \eqref{eq:dae_system} is analogous to conventional DAEs \citep{biegler2010nonlinear}, given a particular set of parameter values $\bof{p}$ and $\bof{\bs{\theta}}$. 
 At this point, we do not make specific restrictions on the index of \eqref{eq:dae_system}. In Sec. \ref{sec:proposed_approach}, we will provide some details on how our approach can deal with higher-index DAEs.

We have access to noisy data collected from a set of trajectories (or scenarios) ${\mca{S}{=}\{1,...,n_s\}}$. Although the output terms $\bof{z}(t)$ of the neural network are usually unobserved, we assume that there are observations for all variables which define the input to the neural network. For the problems considered in this work, this always means the differential state variables. Thus, along each trajectory, observations of the differential states $\bof{x}(t)$ are available at specific times ${t\in\mca{T}^{s}_o}$. Observations of algebraic variables may be available as well; however, we do not consider this here.

In the remainder of this work, we will use the following notation: Vector-valued variables will be noted in lower-case bold face, with a superscript indicating the trajectory the variable is defined on, e.g. $\bof{x}^{(s)}(t)$. Furthermore, discretizations of continuous variables at some index $k$ will be denoted $\bof{x}^{(s)}_k$.
With this, we denote the observed data as:
\begin{align*}
   \hat{\bof{x}}^{(s)}_i = \bar{\bof{x}}^{(s)}(t_i) + \bs{\epsilon}^{(s)}_{i}, 
\quad \forall_{s \in \mca{S}}, \forall_{t_i \in \mca{T}^{s}_o},
\end{align*}
where $\bar{\bof{x}}^{(s)}(t)$ is the \textit{ground truth} trajectory of the underlying system. Unless otherwise stated, we expect the observation noise $\bs{\epsilon}^{(s)}_{i}$ to be drawn from a Gaussian distribution with mean zero. The loss incurred by a continuous state profile $\bof{x}^{(s)}(t)$ with respect to the observed data on trajectory $s$ may be defined as
\begin{equation}\label{eq:cont_data_loss}
\varphi^{(s)}(\bof{x}^{(s)}(t))
= \displaystyle\sum_{t_i\in\mca{T}^{s}_o}
  \left\lVert \bof{x}^{(s)}(t_i) - \hat{\bof{x}}^{(s)}_i\right\rVert_2^2,
\end{equation}
although other formulations (e.g. weighted least squares, absolute error, etc.) are possible.
The general problem formulation for training the neural DAE model is then given by:
\begin{subequations}\label{eq:parmest_opt_NODE}
    \begin{align}
        \displaystyle\min \quad 
        &\displaystyle\sum_{s\in\mca{S}}\varphi^{(s)}(\bof{x}^{(s)}(t))  + \alpha_r r(\bs{\theta}) \label{eq:pNODE_obj}\\
        \text{s. t.} \quad 
        & \frac{d\bof{x}^{(s)}}{dt} = \bof{f}(\bof{x}^{(s)}(t), \bof{y}^{(s)}(t), \bof{z}^{(s)}(t), \bof{p}), 
        &&\forall_{s\in\mca{S}},\ \forall_{t\in[t_0^s, t_f^s]} \label{eq:ode_constr_parmest}\\
        & \bof{h}(\bof{x}^{(s)}(t), \bof{y}^{(s)}(t), \bof{z}^{(s)}(t), \bof{p}) = 0, 
        &&\forall_{s\in\mca{S}},\ \forall_{t\in[t_0^s, t_f^s]}  \label{eq:pNODE_eq_constr}\\
        & \bof{g}(\bof{x}^{(s)}(t), \bof{y}^{(s)}(t), \bof{z}^{(s)}(t), \bof{p}) \le 0, 
        &&\forall_{s\in\mca{S}},\ \forall_{t\in[t_0^s, t_f^s]}  \label{eq:pNODE_ineq_constr}\\
        & \bof{z}^{(s)}(t)  = \bof{f}_{\text{NN}}(\bof{x}^{(s)}(t),\bs{\theta}), 
        &&\forall_{s\in\mca{S}},\ \forall_{t\in[t_0^s, t_f^s]} \label{eq:nn_constraint}\\
        & \bof{x}^{(s)}(t_0^s) = \bof{x}_0^{(s)}(\bof{p}),
        &&\forall_{s\in\mca{S}}. \label{eq:parmest_init}
    \end{align}
\end{subequations}

The parameters of the neural network $\bs{\theta}$ and the static parameters for the mechanistic portion of the model formulation $\bof{p}$ are shared between all trajectories, while each trajectory has its own dynamic state evolution and initial conditions. A general regularization term on the weights of the neural network is included in the objective, with coefficient $\alpha_r$. Unless stated otherwise, we use $r(\bs{\theta}) {=} \frac{1}{2} \norm{\bs{\theta}}_2^2$. Note that the neural DAE \eqref{eq:dae_system} is embedded in the problem, under the addition of inequality, or path constraints \eqref{eq:pNODE_ineq_constr}, which can represent physically-motivated variable bounds, for example. If these bounds become active, the existence of a solution to \eqref{eq:parmest_opt_NODE} might not be guaranteed unless sufficient independent variables are included in the problem. To this end, slack variables and/or barrier reformulations for \eqref{eq:pNODE_ineq_constr} may be used.
Note that the above formulation is still in continuous form, i.e. the discretization of the embedded DAE has not been performed yet. As noted in Sec. \ref{sec:intro}, so-called sequential solution approaches transform \eqref{eq:parmest_opt_NODE} into an unconstrained optimization problem in terms of $\theta$, where \eqref{eq:ode_constr_parmest}-\eqref{eq:parmest_init} are handled through differentiable pipelines involving integration, projection and/or penalty terms. In this work, we present the holistic solution of \eqref{eq:parmest_opt_NODE} using orthogonal collocation and nonlinear programming, i.e., the simultaneous approach for DAE-constrained optimization. Furthermore, we present a bi-level decomposition of the resulting NLP which gives rise to a scalable sequential variant of the proposed approach in Section \ref{sec:decomp}.

\section{Simultaneous approach for neural DAEs} \label{sec:proposed_approach}

We briefly outline the simultaneous approach before applying it to \eqref{eq:parmest_opt_NODE}. At its core, this entails fully discretizing  \eqref{eq:parmest_opt_NODE}, transforming it into a (large-scale) nonlinear optimization problem, which can be solved using an interior point method, for example. Although this approach is agnostic to the discretization scheme, orthogonal collocation is often chosen in practice due to its favorable numerical stability and accuracy.  
In particular, \citet{ascher1998computer} note that collocation methods using Radau points are well-suited  for initial-value DAEs and very stiff ODEs. The application to semi-explicit index-2 DAEs usually necessitates the derivation of consistent initial conditions -- as we will see, this can be directly incorporated in the nonlinear programming problem where the collocation scheme is embedded. For parameter estimation problems, potential non-uniqueness of such conditions can reasonably be expected to be resolved by the minimization of an objective function encoding data fit, if all differential states are observed. Furthermore, for higher-index problems, it has been observed that the simultaneous approach can still return solutions, albeit under some numerical strain stemming from the ill-conditioning of the underlying DAE (\citep{biegler2010nonlinear}, Ch. 10.4).

For the description of the collocation scheme, we omit the superscript $(s)$ denoting the trajectory for notational brevity. The time horizon $[t_0, t_f]$ is divided into $n_{fe}$ finite elements, where element $i$ corresponds to the time span $[t_{i-1}, t_i]$, the length of the element is denoted by $h_i {=} t_i - t_{i-1}$. On each element $i$, the differential states $\bof{x}(t)$ are approximated by polynomials of degree $K$, $\tilde{\bof{x}}(t)$, using Lagrange interpolating polynomials:
\begin{align} \label{eq:collocation}
\begin{split}
    &\tilde{\bof{x}}(t) = \displaystyle\sum_{j=0}^K \ell_j(\tau)\bof{x}_{ij}, \quad t \in [t_{i-1}, t_i], ~~ \tau \in [0,1], \\
    \text{where} \quad &t = t_{i-1} + h_i\tau \quad \text{and} \quad    \ell_j(\tau) = \displaystyle \prod_{k=0, k\neq j}^K \frac{\tau - \tau_k}{\tau_j - \tau_k}.
\end{split}
\end{align}
Here, $\{\tau_j\}_{j=0,...,K}$ are the collocation points on each element, with $\tau_0=0$. The collocation points can be chosen based on different quadrature schemes, e.g. Lagrange-Radau \citep{biegler2010nonlinear}. Here, we assume matching collocation points for each state in $\bof{x}(t)$, but this is not required. Indeed, even the number and size of the finite elements can be adjusted specifically for each differential state. Like all aspects relating to the discretization, this would have to be specified before the resulting NLP is solved. Using the Lagrange polynomials, we have ${\tilde{\bof{x}}(t_{i-1} + \tau_j h_i) {=} \bof{x}_{ij}}$, i. e., the value of the interpolating polynomials $\tilde{\bof{x}}(t)$ at the collocation points is equal to the polynomial coefficients $\bof{x}_{ij}$.

After introducing similar discretizations for the algebraic variables, the continuous problem \eqref{eq:parmest_opt_NODE} is fully discretized:
\begin{subequations}\label{eq:colloc_NODE}
    \begin{align}
        \displaystyle\min \quad &\displaystyle\sum_{s\in\mca{S}}\varphi^{(s)}(\tilde{\bof{x}}^{(s)}(t))  + \alpha_r r(\bs{\theta}) \\
        \text{s. t.} \quad &  \displaystyle\sum_{j=0}^{K} \bof{x}^{(s)}_{ij} \cdot\ell_j'(\tau_k) = h_i \bof{f}(\bof{x}^{(s)}_{ik}, \bof{y}^{(s)}_{ik}, \bof{z}^{(s)}_{ik}, \bof{p}), &&\forall_{s\in\mca{S}}, \forall_{i=1...n_{fe}},\forall_{k=1...K} \label{eq:discr_ndae_ode}\\
        & \bof{h}(\bof{x}^{(s)}_{ik}, \bof{y}^{(s)}_{ik}, \bof{z}^{(s)}_{ik}, \bof{p}) = 0, &&\forall_{s\in\mca{S}}, \forall_{i=1...n_{fe}},\forall_{k=1...K} \label{eq:discr_ndae_eq}\\
        & \bof{g}(\bof{x}^{(s)}_{ik}, \bof{y}^{(s)}_{ik}, \bof{z}^{(s)}_{ik}, \bof{p}) \le 0 , &&\forall_{s\in\mca{S}}, \forall_{i=1...n_{fe}},\forall_{k=1...K}\label{eq:discr_ndae_ineq}\\
        &\bof{z}^{(s)}_{ik} = \bof{f}_{\text{NN}}(\bof{x}^{(s)}_{ik},\bs{\theta}), &&\forall_{s\in\mca{S}}, \forall_{i=1...n_{fe}},\forall_{k=1...K} \label{eq:discr_nn_constraint}\\
        & \bof{x}^{(s)}_{i+1, 0} = \sum_{j=0}^K \ell_j(1)\bof{x}^{(s)}_{ij}, &&\forall_{s\in\mca{S}}, \forall_{i=1...n_{fe}},\forall_{k=1...K}\label{eq:discr_cont_constraint}\\
        & \bof{x}^{(s)}_{1,0} = \bof{x}^{(s)}_0(\bof{p}). \label{eq:discr_ndae_init}
    \end{align}
\end{subequations}
 Here, the continuity of the state variables is enforced through \eqref{eq:discr_cont_constraint}, similar constraints can be defined for the algebraic variables, if desired. Note that the continuous state profiles (the collocation polynomials) $\tilde{\bof{x}}^{(s)}(t)$ can be evaluated at arbitrary times in the objective, i.e., the discretization points do not have to match the points where data were observed.
 
 We emphasize that $\bof{x}_{ik}, \bof{y}_{ik}, \bof{z}_{ik}$, $\bs{\theta}$ and $\bof{p}$ are variables in Problem \eqref{eq:colloc_NODE}. Therefore, this problem is significantly more challenging to solve than conventional DAE-constrained optimization problems, due to the potentially large number of additional variables (the weights and biases of the neural network) and the additional constraints \eqref{eq:discr_nn_constraint}, which are highly nonconvex and dense in $\bs{\theta}$. Solving \eqref{eq:colloc_NODE} directly using a nonlinear programming solver often proved computationally intractable. Thus, in order to obtain high-quality, locally optimal solutions to \eqref{eq:colloc_NODE}, we apply a number of pre-processing steps, which are outlined in the following sections.

It is important to note that the formulation of the NLP \eqref{eq:colloc_NODE} and its solution with an interior point method makes certain restrictions on the type of neural network we can consider. Namely, we require the function $\bof{f}_{\text{NN}}$ to have smooth second-order derivatives. In this work, we will exclusively consider dense feed-forward neural networks with smooth activation functions, e.g. `sigmoid', `softplus', or similar. As will be discussed in Section \ref{sec:ipm_NODE}, it is not required to be able generate algebraic expressions encoding the neural network. An interface to evaluate the function, its Jacobian, and optionally its Hessian is sufficient. Furthermore, normalization layers for the input and output of the neural network proved to aid the robustness of our approach. The associated constants can be computed from the initialization procedure outlined in Section \ref{sec:initialization}.

\subsection{Initialization strategy}\label{sec:initialization}

In order to make the solution of \eqref{eq:colloc_NODE} more tractable, we introduce an auxiliary problem, which is solved to obtain initial estimates for the trajectories of the differential states $\bof{x(t)}$ and algebraic variables $\bof{z}(t), \bof{y}(t)$. To this end, constraint \eqref{eq:discr_nn_constraint} and the variables associated with the neural network, $\bs{\theta}$, are removed from problem \eqref{eq:colloc_NODE}. The independent variables of this problem are the discretized unknown algebraic terms $\bof{z}^{(s)}_{ik}$ for each trajectory $s$, which are chosen to minimize the same observation loss with respect to the observed data as before, augmented with a smoothness penalty for the polynomial $\tilde{\bof{z}}(t)$ on each finite element:

\begin{subequations}\label{eq:smooth_init}
\begin{align}
    \min &\displaystyle\sum_{s\in\mca{S}} \left[\varphi^{(s)}(\tilde{\bof{x}}^{(s)}(t))  + \alpha_{s}\sum_{i=1}^{n_{fe}}\sum_{k, j=1}^K ||\bof{z}_{ij}\cdot \ell_j'(\tau_k)||_2^2\right] \\
    \text{s.t.} \quad &\text{\eqref{eq:discr_ndae_ode}}, \text{\eqref{eq:discr_ndae_eq}}, \text{\eqref{eq:discr_ndae_ineq}}, \text{\eqref{eq:discr_cont_constraint}}, \text{\eqref{eq:discr_ndae_init}}
\end{align}
\end{subequations}

By solving Problem \eqref{eq:smooth_init}, we obtain trajectories for the differential and algebraic variables, which we denote by $\bof{x}^{(s)}_{\text{init}}(t)$ and $\bof{z}^{(s)}_{\text{init}}(t), \bof{y}^{(s)}_{\text{init}}(t)$, respectively. We also obtain values for the static variables, $\bof{p}_{\text{init}}$. They adhere to the constraints defined for the dynamic optimization problem at hand, at the discretization points defined by the collocation scheme.

Problem \eqref{eq:smooth_init} is computationally tractable, compared to \eqref{eq:colloc_NODE}, as it contains fewer variables and omits the highly nonconvex constraints associated with the neural network. Furthermore, it is perfectly separable by trajectory, so the solution can be trivially parallelized. If there are static variables $\bof{p}$ that must match across trajectories, their initial value can be obtained by averaging the values obtained from the solution of each subproblem.

We use the solution of \eqref{eq:smooth_init} for three purposes related to the initialization of variables in \eqref{eq:colloc_NODE}. For each trajectory, we initialize $\bof{x}^{(s)}_{ik}$, $\bof{z}^{(s)}_{ik}$ and $\bof{y}^{(s)}_{ik}$ by evaluating $\bof{x}^{(s)}_{\text{init}}(t)$, $\bof{z}^{(s)}_{\text{init}}(t)$, and $\bof{y}^{(s)}_{\text{init}}(t)$ at the appropriate discretization points, respectively. Secondly, the input and output normalization layers of the neural network are fixed by computing the mean/variance of the discretized trajectories obtained from \eqref{eq:smooth_init}. Finally, we can obtain initial values for $\bs{\theta}$ by running stochastic gradient descent (SGD), with a fixed number of epochs and batch size, on a loss function defined by the trajectories for the input and output of the neural network, i.e. $\bof{x}^{(s)}_{\text{init}}(t)$ and $\bof{z}^{(s)}_{\text{init}}(t)$, respectively;
\begin{align}\label{eq:nn_init}
\varphi_{\text{init}}(\bs{\theta}) = \sum_{s\in\mca{S}}\sum_{t_i \in \mca{T}^s_{\text{init}}} \left(\bof{z}^{(s)}_{\text{init}}(t_i) - \bof{f}_{NN}(\bof{x}^{(s)}_{\text{init}}(t_i), \bs{\theta})\right)^2.
\end{align}
The evaluation points $\mca{T}^{s}_{\text{init}}$ for this step can be chosen arbitrarily; however, it is sensible to coordinate them with the collocation scheme used to solve \eqref{eq:colloc_NODE}.
At this point, we have obtained initial values for all relevant variables in \eqref{eq:colloc_NODE}, i. e., trajectories of the differential and algebraic variables, as well as the weights of the neural network. Note that the use of this initialization scheme may introduce biases towards learned representations that exhibit some degree of smoothness, which must be weighed against the potential improvements to the robustness and convergence of the solution of \eqref{eq:colloc_NODE}.

In the next section, we describe some of the computational challenges encountered when solving \eqref{eq:colloc_NODE} using an interior point method.

\subsection{Computational aspects of applying an interior point method to NLPs with neural network components}\label{sec:ipm_NODE}
In this section, we discuss computational aspects of solving problems similar to \eqref{eq:colloc_NODE}, i.e., constrained, nonconvex, nonlinear optimization problems, where a subset of the variables and constraints is defined by a neural network. Since we use \textsc{IPOPT} throughout our approach, we follow the interior point method (IPM) as described in \citet{wachter2006}, highlighting the challenges arising from the incorporation of neural networks. We will omit many specifics, and refer to \citet{wachter2006} for an in-depth discussion of the algorithm.
Within the interior point method, a series of barrier subproblems, where inequality terms are moved to the objective, are solved while subsequently decreasing the barrier coefficient $\mu$. For the problem \eqref{eq:colloc_NODE}, a general formulation of the barrier subproblem can be denoted by
\begin{subequations}\label{eq:general_nlp_w_theta}
    \begin{align}
    \displaystyle\min_{\bs{\nu}, \bs{\theta}} \quad & f(\bs{\nu}) + \alpha_r r(\bs{\theta}) - \mu \sum_i \ln(\bof{s}_i)\\
    \text{s. t.} \quad & \bof{c}(\bs{\nu}) = 0 &&[\bs{\lambda}]\\
    & \bof{d}(\bs{\nu}, \bs{\theta}) = 0 &&[\bs{\rho}] \label{eq:general_nlp_nn_constr}\\
    & \bof{g}(\bs{\nu}) + \bof{s} = 0 &&[\bof{z}]
    \end{align}
\end{subequations}
where $\bs{\theta}$ are the weights and biases of the neural network, and $\bs{\nu}$ contains all other primal variables, i. e. ${\bs{\nu} = (\cdots,  \bof{x}_{ik}, \bof{y}_{ik}, \bof{z}_{ik}, \cdots, \bof{p})^{\top}}$, for all discretization points $ik$. Slightly overloading our notation, we collect all constraints from \eqref{eq:colloc_NODE} into $\bof{c}(\bs{\nu})$, $\bof{g}(\bs{\nu})$ and $\bof{d}(\bs{\nu}, \bs{\theta})$, where the latter contains the neural network expressions \eqref{eq:discr_nn_constraint}. We denote the constraint multipliers in square brackets. The data fit component of the objective function is given by ${f(\bs{\nu}) = \sum_{s\in\mca{S}}\varphi^{(s)}(\tilde{\bof{x}}^{(s)}(t))}$. The corresponding Lagrangian is defined as
\begin{equation}
    \mca{L}(\bs{\nu}, \bs{\theta}, \bs{\lambda},\bs{\rho}, \bof{z}) = f(\bs{\nu}) + \alpha_r r(\bs{\theta}) - \mu \sum_i \ln(\bof{s}_i) + \bs{\lambda}^{\top}\bof{c}(\bof{x}) + \bs{\rho}^{\top}\bof{h}(\bof{x}, \bs{\theta}) + \bof{z}^{\top}(\bof{g}(\bs{\nu}) + \bof{s}).
\end{equation}

The interior-point method solves the KKT conditions of \eqref{eq:general_nlp_w_theta} using a Newton-type method. At every iteration, this produces a linear system of equations, which must be solved to obtain a step direction:
\begin{align}\label{eq:ipm_step}
\underbrace{
    \bbm 
    \bof{W}_{\nu\nu} + \sigma_H \bof{I} & \bof{W}_{\theta\nu}^{\top} & \nabla_\nu \bof{c}^{\top} & \nabla_\nu \bof{d}^{\top} & \nabla_{\nu} \bof{g}^{\top}\\[1mm]
    \bof{W}_{\theta\nu} & \bof{W}_{\theta\theta} + \sigma_H \bof{I} & 0 & \nabla_\theta \bof{d}^{\top} & 0\\[1mm]
    \nabla_\nu \bof{c} & 0 & -\sigma_C \bof{I} & 0 & 0\\[1mm]
    \nabla_\nu \bof{d} & \nabla_\theta \bof{d} & 0 &  -\sigma_C \bof{I} & 0\\
    \nabla_{\nu} \bof{g} & 0 & 0 & 0 & -\bs{\Sigma}^{-1}
    \ebm}_{\bof{A}}
    \bbm
    \bof{b}_\nu \\[.8mm]
    \bof{b}_\theta \\[.8mm]
    \bof{b}_\lambda \\[.8mm]
    \bof{b}_\rho \\[.8mm]
    \bof{b}_z
    \ebm
     &=
    -
    \begin{bmatrix}
        \nabla_{\nu} \mca{L} \\[.8mm]
        \nabla_{\theta} \mca{L} \\[.8mm]
        \bof{c}(\bs{\nu})\\[.8mm]
        \bof{d}(\bs{\nu}, \bs{\theta})\\[.8mm]
        \bof{g}(\bs{\nu}) + \mu\bof{Z}^{-1}\bof{e}
    \end{bmatrix},
\end{align}
where $\bs{\Sigma} {=} \bof{S}^{-1}\bof{Z}$ and ${\bof{e} {=} (1,1 \ldots, 1)^{\top}}$.
Note that $\delta_H$ and $\delta_C$ are scalar terms which are adjusted to correct the inertia of $\bof{A}$, to ensure that the resulting step direction achieves descent.
The Hessian terms are given by
\begin{subequations}\label{eq:hess_terms}
\begin{align}
    &\bof{W}_{\nu\nu} = \nabla^2_{\nu\nu} f(\bs{\nu}, \bs{\theta}) + \bs{\lambda}^{\top}\nabla^2_{\nu\nu} \bof{c}(\bs{\nu}) + \bs{\rho}^{\top}\nabla^2_{\nu\nu} \bof{d}(\bs{\nu}, \bs{\theta}) + \bof{z}^{\top}\nabla^2_{\nu\nu} \bof{g}(\bof{x}, \bs{\theta})\\
    &\bof{W}_{\theta\nu} = \bs{\rho}^{\top}\nabla^2_{\theta\nu} \bof{d}(\bof{x}, \bs{\theta})\\
    &\bof{W}_{\theta\theta} = \alpha_r \bof{I} + \bs{\rho}^{\top}\nabla^2_{\theta\theta} \bof{d}(\bs{\nu}, \bs{\theta}).
\end{align}
\end{subequations}

At every step of the IPM, \eqref{eq:ipm_step} must be evaluated and solved, which poses the following computational challenges for the problem at hand:
\begin{enumerate}
    \item Hessian and Jacobian terms involving $\bs{\theta}$, i. e., $\bof{W}_{\theta\nu}$, $\bof{W}_{\theta\theta}$ and $\nabla_{\theta} \bof{d}$ are dense, owing to the architecture of feed-forward neural networks. This makes them expensive to evaluate using modeling tools such as the \textsc{AMPL} solver library \citep{fourer1990modeling}, which is commonly used to generate expressions and evaluate derivatives as part of NLP solver interfaces.
    \item The solution of \eqref{eq:ipm_step} using sparse indefinite solvers such as HSL MA27 \citep{duff1982ma27} becomes more computationally expensive, both due to the increased density and the frequent need for regularization and re-factorization stemming from the degeneracy of $\bof{d}(\bs{\nu}, \bs{\theta})$ in both $\bs{\nu}$ and $\bs{\theta}$.
\end{enumerate}

Existing tools can be used to address the first point, namely so-called external gray-box modeling interfaces (e. g. for the \textsc{Pyomo} modeling language\footnote{\url{https://pyomo.readthedocs.io/en/6.4.3/contributed_packages/pynumero/pynumero.interfaces.external_grey_box_model.html}}). Thereby, the constraints involving the neural network \eqref{eq:general_nlp_nn_constr} are treated as an oracle whose value, Jacobian and Hessian can be evaluated. These evaluations can then be performed using modern deep learning toolkits such as JAX \citep{jax2018github}, which are are specialized for these operations on neural networks. Furthermore, the specific structure of $\bof{d}(\bs{\nu}, \bs{\theta})$ for the problem at hand, i.e.
\begin{equation}
    \bof{d}(\bs{\nu}, \bs{\theta}) =
    \bbm
    \vdots\\
    \bof{z}^{(s)}_{ik} - \bof{f}_{\text{NN}}(\bof{x}^{(s)}_{ik},\bs{\theta})\\
    \vdots
    \ebm, 
\end{equation}
allows for the vectorized evaluation of the constraints and associated Jacobians/Hessians across all discretization points ${\{ik\}_{ \forall_{i=1...n_{fe},k=1...K}}}$ and trajectories $\{(s)\}_{\forall_{s\in\mca{S}}}$ defined in \eqref{eq:colloc_NODE}. It should be noted that even when using this approach, the evaluation of $\bof{W}_{\theta\theta}$ can become expensive for large neural networks.

To sidestep this potential bottleneck and to deal with the aforementioned challenges associated with the solution of \eqref{eq:ipm_step}, an L-BFGS approximation for the Hessian terms can be used. This is a readily available option in solvers such as \textsc{IPOPT} and involves collecting iterate samples as the interior point method progresses to form matrices $\bof{B}$ and $\bof{M}$, which define a Hessian approximation $\tilde{\bof{H}} {=} \xi\bof{I} + \bof{B}\bof{M}\bof{B}^{\top}$ (see \citet{nocedal1999numerical}, Sec. 7.2 for details). Thus, $\bof{A}$ in \eqref{eq:ipm_step} can be replaced by
\begin{align}\label{eq:lbfgs_step}
     \bbm 
   \xi \bof{I}  & 0 & \nabla_\nu \bof{c}^{\top} & \nabla_\nu \bof{h}^{\top} & \nabla_{\nu} \bof{g}^{\top}\\[1mm]
    0 & \xi \bof{I}  & 0 & \nabla_\theta \bof{h}^{\top} & 0\\[1mm]
    \nabla_\nu \bof{c} & 0 & -\sigma_C \bof{I} & 0 & 0\\[1mm]
    \nabla_\nu \bof{h} & \nabla_\theta \bof{h} & 0 &  -\sigma_C \bof{I} & 0\\
    \nabla_{\nu} \bof{g} & 0 & 0 & 0 & -\bs{\Sigma}^{-1}
    \ebm + 
    \bbm 
      \\
    \bof{B}\\[2mm]
    0 \\
    0\\
    0
    \ebm
    \bbm
   &\bof{M}\bof{B}^{\top} & 0 & 0 & 0
    \ebm.
\end{align}
To compute a solution to the resulting linear system, only the factorization of the left matrix in \eqref{eq:lbfgs_step} is necessary (\citet{nocedal1999numerical}, Sec. 19.3), which is significantly cheaper than solving \eqref{eq:ipm_step}, due to increased sparsity and the fact that $\tilde{\bof{H}}$ is guaranteed to be positive definite, thus removing the need for curvature correction. However, because of the use of the Hessian approximation, the quality of the resulting step is often inferior, leading to an increased number of IPM iterations to converge to a solution.
For the problem considered in this work \eqref{eq:colloc_NODE}, the use of the L-BFGS approximation proved highly effective. This is likely due to a combination of the factors described above (density, nonconvexity and Hessian evaluations). Similar results using the L-BFGS approximation were recently observed by \citet{parker2024formulations} when embedding trained neural networks into optimization problems.
One potential drawback of the Hessian approximation strategy is that it replaces the full Hessian by an approximation. It seems more promising to only approximate the components of the Hessian which cause computational difficulties, i.e. those associated with the neural network. This would retain the exact Hessians of the mechanistic part of the problem thus likely improving the quality of the resulting step directions. This is an interesting avenue for future research.

In Section \ref{sec:case_studies}, we will show that the strategies described above, in combination with the initialization scheme from Section \ref{sec:initialization}, make the simultaneous approach for training neural DAEs tractable for problems of small to medium scale, in terms of the size of the neural network and the number of trajectories considered.
However, the approach outlined so far has a straightforward scalability issue: The size of the NLP directly scales with number of parameters in the neural network and the number of trajectories included in the training problem \eqref{eq:colloc_NODE}. Even when using the L-BFGS approximation and gray-box modeling components, the size and number of nonzeros in \eqref{eq:lbfgs_step} increases accordingly, resulting in more costly factorizations at every step of the interior point method. Thus, to scale the approach to large-scale learning problems, decomposition approaches need to be applied.

\subsection{Bi-level decomposition using NLP sensitivities} \label{sec:decomp}

Problem \eqref{eq:colloc_NODE} naturally lends itself to well-established (parallel) decomposition schemes for NLPs with block structure, where the a set of subproblems corresponding to the separate trajectories are linked by the common variables $\bs{\theta}$ and $\bof{p}$ (we will disregard $\bof{p}$ in the following discussion in the interest of notational clarity). These kinds of structures are commonly exploited using Schur-complement approaches \citep{zavala2008interior, kang2014interior} or problem-level decompositions like the alternating direction  method of multipliers (ADMM) \citep{shapovalova2025training}. The latter does not guarantee convergence to a local solution for nonconvex problems, but is known to achieve good results in practice. These types of decomposition would remedy the scaling issue of the simultaneous approach for neural DAEs with respect to the number of trajectories. However, as the size of the neural network, and thereby the number of linking variables $\bs{\theta}$, grows, the parallel efficiency of Schur-complement-type decompositions is known to deteriorate \citep{lueg2025domain}. In the case of ADMM, each subproblem would still contain the linking variables, so the unfavorable scaling properties of the simultaneous approach are not expected to change.

Instead, we propose a bi-level decomposition which is conceptually similar to the sequential approaches for training neural DAEs mentioned in Section \ref{sec:intro}: the outer level updates the neural network parameters using (stochastic) gradient descent, where gradients are computed by evaluating the sensitivity of an inner problem, which corresponds to \eqref{eq:colloc_NODE} with $\bs{\theta}$ fixed. This produces much more tractable NLPs which can be trivially separated by trajectory and still benefit from the gray-box functionality to evaluate derivatives of the neural network with respect to its inputs. Furthermore, the approach leverages the initialization scheme outlined in Section \ref{sec:initialization}, to reduce the number of gradient steps required on the outer level to converge to a satisfactory solution. Clearly, this approach will not provably converge to a locally optimal solution of \eqref{eq:colloc_NODE} -- however, it can ensure strict constraint satisfaction, which is of particular interest in the area of learning hybrid models. In practice, gradient descent is usually terminated once the objective stops improving significantly.

We will show that the evaluation of the sensitivity of the subproblem solutions with respect to $\bs{\theta}$ can be evaluated cheaply, again leveraging software tools geared towards automatic differentiation for deep learning models. 

The outer problem is defined as
\begin{align}\label{eq:decomp_UP}
    \min_{\bs{\theta}} \quad 
    \Phi(\bs{\theta}) 
    := \alpha_r r(\bs{\theta}) + \displaystyle\sum_{s\in\mca{S}} \tilde{\varphi}^{(s)}(\tilde{\bs{\nu}}^{(s)}(\bs{\theta})), \quad \text{s. t.}\quad \tilde{\bs{\nu}}^{(s)}(\bs{\theta}) \in \arg\min ~\mathcal{S}^{(s)}_{\theta},
\end{align}
where each subproblem $\mathcal{S}^{(s)}_{\theta}$ corresponds to \eqref{eq:colloc_NODE} for a single trajectory $s$ with $\bs{\theta}$ fixed. For the remainder of this section, we will drop the superscript indicating the trajectory. Clearly, the subproblems can be solved in parallel and their sensitivities are summed to compute the gradient of $\Phi$ \eqref{eq:decomp_UP}. With this, the subproblem is denoted as
\begin{subequations}\label{eq:decomp_SP}
\begin{align}
    \mathcal{S}_{\theta}:\quad
    \min_{\tilde{\bs{\nu}}, \tilde{\bof{s}}} \quad 
    & \tilde{\varphi}(\tilde{\bs{\nu}}) - \mu \sum_i \ln(\tilde{\bof{s}}_i) &&\\
    \text{s.t.} \quad 
    & \bof{c}(\tilde{\bs{\nu}}) = 0, && [\bs{\lambda}]\\
    & \bof{d}_{\theta}(\tilde{\bs{\nu}}) = 0, &&[\bs{\rho}] \label{eq:decomp_sp_nn_constr}\\
    &\tilde{\bof{g}}(\tilde{\bs{\nu}}) + \tilde{\bof{s}} = 0.  &&[\bof{z}]
\end{align}
\end{subequations}
This is similar to the general formulation \eqref{eq:general_nlp_w_theta} introduced in Section \ref{sec:proposed_approach}, with the following adaptations:
\begin{align} \label{eq:decomp_SP_defs}
    &\tilde{\bs{\nu}} = \begin{bmatrix}
        \bs{\nu}, \\\vdots \\\bs{\Delta}^{+}_{ik} \\ \bs{\Delta}^{-}_{ik} \\ \vdots
    \end{bmatrix}, \quad
    \bof{d}_{\theta}(\tilde{\bs{\nu}}) = 
    \begin{bmatrix}
        \vdots \\
         \bof{z}_{ik} - \bof{f}_{\text{NN}}(\bof{x}_{ik}, \bs{\theta}) - \bs{\Delta}^{+}_{ik} + \bs{\Delta}^{-}_{ik} \\
        \vdots
    \end{bmatrix}, \quad 
    \tilde{\bof{g}}(\tilde{\bs{\nu}}) = 
    \begin{bmatrix}
        \bof{g}(\bs{\nu}) \\
        \vdots \\
         - \bs{\Delta}^{+}_{ik}\\
         - \bs{\Delta}^{-}_{ik}\\
        \vdots
    \end{bmatrix},\\[2mm]
    &\tilde{\varphi}(\bof{\tilde{\bs{\nu}}}) = \varphi(\tilde{\bof{x}}(t)) + \alpha \bof{e}^{\top}\sum_{ik}(\bs{\Delta}^{+}_{ik} + \bs{\Delta}^{-}_{ik}).
\end{align}
We use the vertical dots to indicate that the constraints and variables are defined over all discretization points $ik$. The only significant changes are that $\bs{\theta}$ is no longer a variable and that non-negative slack variables $\bs{\Delta}$ for the neural network constraints were introduced, to ensure that a feasible solution to \eqref{eq:decomp_SP} exists. The objective is modified to penalize the absolute value of these slacks -- given that the coefficient $\alpha$ is sufficiently large, this ensures that a solution with zero slack is found, if it exists. This ensures strict constraint satisfaction as the algorithm progresses. Note that we still include the data objective in $\tilde{\varphi}$, as this proved to aid the robustness of the solution of \eqref{eq:decomp_SP}.

When applying gradient descent to \eqref{eq:decomp_UP} we note that
\begin{align}\label{eq:grad_upper_loss}
    \frac{d \Phi}{d\bs{\theta}} = \alpha_r \frac{d r}{d\bs{\theta}} + \nabla_{\bs{\theta}}\tilde{\bs{\nu}}(\bs{\theta})^{\top}\frac{d \tilde{\varphi}}{d \tilde{\bs{\nu}}}.
\end{align}
To compute this gradient, we need to be able to evaluate a vector product with the sensitivity of the solution of $\mathcal{S}^{\mu}_{\theta}$ with respect to $\bs{\theta}$. At the solution of \eqref{eq:decomp_SP}, the KKT conditions are satisfied:
\begin{align}\label{eq:kkt_SP}
    \bof{F}(\bof{q}(\bs{\theta}), \bs{\theta}) =
    \begin{bmatrix}
        \nabla_{\tilde{\nu}} \tilde{\varphi} + \nabla_{\tilde{\nu}}\bof{c}^{\top}\bs{\lambda} + \nabla_{\tilde{\nu}}\bof{d}_{\theta}^{\top}\bs{\rho} + \nabla_{\tilde{\nu}} \tilde{\bof{g}}^{\top}\bof{z} \\
        \bof{c}(\tilde{\bs{\nu}}) \\
        \bof{d}_{\theta}(\tilde{\bs{\nu}})\\
        \tilde{\bof{g}}(\tilde{\bs{\nu}}) + \tilde{\bof{s}}\\
        \tilde{\bof{S}}\bof{Z} - \mu \bof{e}
    \end{bmatrix} = 0,
\end{align}
where $\bof{q}(\bs{\theta}) {=} (\tilde{\bs{\nu}}(\bs{\theta}), \tilde{\bof{s}}(\bs{\theta}), \bs{\lambda}(\bs{\theta}), \bs{\rho}(\bs{\theta}), \bof{z}(\bs{\theta}))^{\top} \in \mathbb{R}^{n_q}$ is the primal-dual solution of \eqref{eq:decomp_SP}. Using the implicit function theorem, the following relation holds \citep{pacaud2025sensitivity}:
\begin{align}
    &\nabla_{\theta} \bof{q}(\bs{\theta}) = - (\nabla_{q}\bof{F}(\bof{q}, \bs{\theta}))^{-1} \nabla_{\theta}\bof{F}(\bof{q}, \bs{\theta)}\\
    \Rightarrow &\nabla_{\theta} \bof{q}(\bs{\theta})^{\top}\bof{u} = - \nabla_{\theta}\bof{F}(\bof{q}, \bs{\theta)}^{\top}\bar{\bof{u}}, \quad \text{where} \quad \bar{\bof{u}} = (\nabla_{q}\bof{F}(\bof{q}, \bs{\theta}))^{-\top}\bof{u}  \in \mathbb{R}^{n_q}.\label{eq:vjp_kkt_inner}
\end{align}
In order for the resulting sensitivities to be well-defined, we need to ensure that \eqref{eq:decomp_SP} is sufficiently regular, i.e., that linear independence constraint qualifications (LICQ) and strong second-order sufficiency conditions (SSOSC) hold. When solving \eqref{eq:decomp_SP} to local optimality using \textsc{IPOPT} and given that no inertia correction was necessary at the solution, these conditions are met \citep{pirnay2012optimal}. Note that we solve \eqref{eq:decomp_SP} with exact Hessian information, so we can verify this at the solution of every subproblem.

To evaluate \eqref{eq:grad_upper_loss}, we can use \eqref{eq:vjp_kkt_inner} by setting 
${\bof{u} {=} \begin{bmatrix} \frac{d \tilde{\varphi}}{d \tilde{\bs{\nu}}}^{\top}, & 0, & 0, & 0, & 0
\end{bmatrix}^{\top} \in \mathbb{R}^{n_q}}$
and similarly selecting the component corresponding to $\tilde{\bs{\nu}}$ from the resulting vector. 
When solving \eqref{eq:decomp_SP} using an interior point method, $\nabla_{q}\bof{F}(\bof{q}, \bs{\theta})$ is the linear system factorized at every step of the algorithm -- this corresponds to $\bof{A}$ from \eqref{eq:ipm_step} with the second row and column removed, since $\bs{\theta}$ is no longer a variable. A factorization of this matrix could be directly obtained from the solver at the solution, in our case we re-factorize it as we do not have direct access to the solver internals.

Evaluating $\nabla_{\theta}\bof{F}(\bof{q}, \bs{\theta)}$ can be achieved at low computational cost for problems such as \eqref{eq:decomp_SP}.
From \eqref{eq:kkt_SP}, we have
\begin{align}
    &\nabla_{\theta}\bof{F}^{\top} = 
    \begin{bmatrix}
        \nabla_{\theta}(\nabla_{\tilde{\nu}} \bof{d}_{\theta}^{\top}\bs{\rho})^{\top}, & 0, & \nabla_{\theta}\bof{d}_{\theta}^{\top}, & 0, & 0
    \end{bmatrix} \in \mathbb{R}^{n_{\theta} \times n_{q}}, \\
    \Rightarrow &\nabla_{\theta}\bof{F}^{\top}\bar{\bof{u}} = \nabla_{\theta}(\nabla_{\tilde{\nu}} \bof{d}_{\theta}^{\top}\bs{\rho})^{\top}\bar{\bof{u}}_{\tilde{\nu}} + \nabla_{\theta}\bof{d}_{\theta}^{\top}\bar{\bof{u}}_{\rho} \in \mathbb{R}^{n_{\theta}}, \label{eq:vjp_f_theta}
\end{align}
where we indicated particular components of $\bar{\bof{u}}$ by subscripts indicating the corresponding component in $\bof{q}$.
Using the definition of $\bof{d}_{\theta}$ from \eqref{eq:decomp_SP_defs}, we get
\begin{align}
\nabla_{\theta}(\nabla_{\tilde{\nu}} \bof{d}_{\theta}^{\top}\bs{\rho})^{\top}\bar{\bof{u}}_{\nu} &= 
 \sum_{ik} -\nabla_{\theta}\left(\nabla_{x} \bof{f}_{\text{NN}}(\bof{x}_{ik}, \bs{\theta})^{\top}\bs{\rho}_{ik}\right)^{\top}\bar{\bof{u}}_{x_{ik}}, 
\label{eq:vjp_stat}\\
\nabla_{\theta}\bof{d}_{\theta}^{\top}\bar{\bof{u}}_{\rho} &= 
\sum_{ik}
 -\nabla_{\theta}\bof{f}_{\text{NN}}(\bof{x}_{ik}, \bs{\theta})^{\top}\bar{\bof{u}}_{\rho_{ik}}. \label{eq:vjp_h}
\end{align}
Here, $\bs{\rho}_{ik}$ indicates the multiplier of constraint \eqref{eq:decomp_sp_nn_constr} at discretization point $ik$. Note that the evaluation of the full vector-Jacobian product (VJP) \eqref{eq:vjp_f_theta} can be achieved by parallel (across discretization points $ik$) evaluations of VJPs of the neural network. This can be implemented with powerful automatic differentiation packages for neural networks, e.g. JAX. In \eqref{eq:vjp_stat} there are nested VJPs, however the inner evaluation only involves low-dimensional cotangent vectors (output dimension of the neural network).

Hence, the evaluation of the sensitivities is expected to be significantly cheaper than the solution of \eqref{eq:decomp_SP}, which likely incurs the majority of the computational cost for each iteration of gradient descent. However, the solution time of \eqref{eq:decomp_SP} is not expected to increase significantly with larger neural networks, as they are contained in a gray-box model whose derivatives can be evaluated cheaply. The changes to the subproblems between iterations of gradient descent are expected to be minor, given appropriate outer step sizes, thus effective warm-starting can be implemented. Lastly, the subproblems corresponding to a particular (groups of) trajectories can be solved in parallel, given appropriate computing infrastructure. In this case, this means distributing the subproblems among the available central processing units (CPU), and ensuring the availability of sufficient memory to run \textsc{IPOPT} on each parallel rank.

\section{Computational experiments} \label{sec:case_studies}

The following case studies analyze some of the capabilities of our proposed methods for different problems of the type described by \eqref{eq:parmest_opt_NODE}. This includes the training procedure, as well as the performance of the trained models on inference tasks. The topic of inference merits some additional discussion here. Once a training procedure is completed, the learned relationship $\bof{f}_{NN}$ or the entire neural DAE can be used for downstream tasks such as simulation, control or other optimization tasks. In our experiments, we will evaluate the accuracy of the learned neural DAE on unseen trajectories (i. e., initial conditions) of the same system. As we have access to the ground truth models, accuracy can be evaluated on the full (state) trajectory of the neural DAE. The accuracy of the learned component can also be evaluated by directly comparing it to the ground truth relationship it was meant to learn, without solving the DAE.

In this work, we have outlined two basic approaches to training neural DAEs using the simultaneous approach: the solution of a `fully' simultaneous NLP \eqref{eq:colloc_NODE}, where the neural network parameters are directly embedded as variables, and the bi-level decomposition approach from Section \ref{sec:decomp}, where a series of more tractable NLPs are solved while updating the neural network parameters using gradient descent. The initialization of both approaches can be refined by leveraging a smooth initialization problem, optionally in combination with the pretraining of the neural network, as outlined in Section \ref{sec:initialization}.

We begin by quantifying the contribution that the initialization schemes and Hessian approximation make towards the tractability and accuracy of the fully simultaneous training procedure in Section \ref{sec:4tank_ablation}. Then, we will investigate the scalability of both methods with respect to the size of the neural networks and the amount of training data in Sections \ref{sec:scalability_decomp}-\ref{sec:scalability_simul}. Lastly, we demonstrate some of the advantages of the simultaneous approach for training (cf. Section \ref{sec:pop_dyn}) and inference (cf. Section \ref{sec:fedbatch}) for hybrid dynamic models with physically motivated (inequality) constraints.

The experiments were run on a conventional laptop, equipped with a 12th Gen Intel(R) Core(TM) i7-12700H (14 cores) and 32 GB of RAM. Scalability tests were performed on the Pittsburgh Supercomputing Center Bridges-2 system \citep{brown2021bridges}, using a single node equipped with two AMD EPYC 7742 CPUs, each with 64 cores, and 256 GB of RAM. The code used to generate the results shown here is made available on GitHub\footnote{\url{https://github.com/llueg/SiNDAE}}.

\subsection{Tank-manifold system} \label{sec:tank}

\begin{figure}
    \centering
    \includegraphics[width=0.4\linewidth]{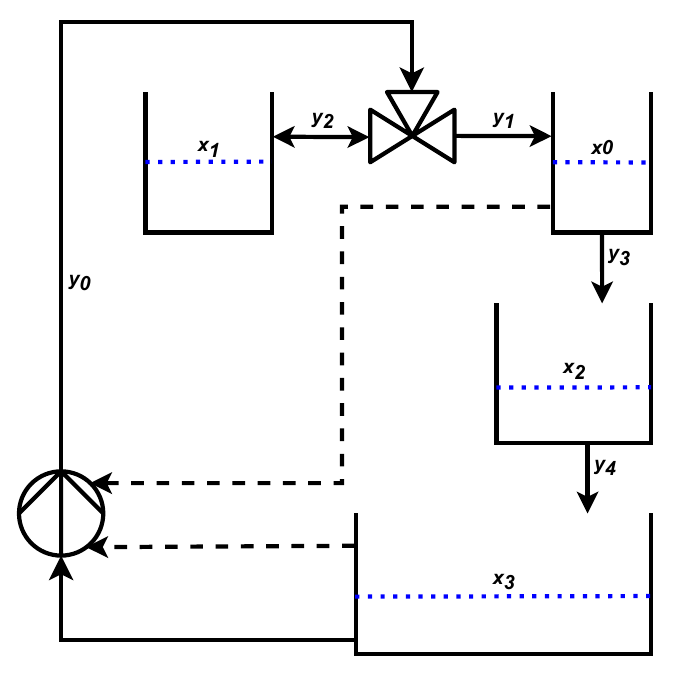}
    \caption{Tank-Manifold system adapted from \citet{koch2024neural}.}
    \label{fig:4tanks_sys}
\end{figure}

We adapt an example DAE system from \citet{koch2024neural}, which describes the dynamics of a closed system of four connected tanks. The differential states $x_0,..,x_3$ describe the fluid heights in the different tanks, the algebraic variables $y_0,..,y_4$ denote the flow rate of liquid between tanks. A pump sets the flow rate $y_0$ as a function of fluid heights $x_0$ and $x_3$. The system is illustrated in Fig. \ref{fig:4tanks_sys}. Importantly, the flow in $y_2$ is reversible, and the fluid heights $x_0$ and $x_1$ are constrained to be equal at all times. The differential-algebraic equations describing the system are given below:
\begin{subequations}\label{eq:tank_dae}
    \begin{align}
        &\frac{d x_0}{dt} = \frac{1}{\phi_0(x_0)} \left( y_1 - y_3\right),\\
        &\frac{d x_1}{dt} = \frac{1}{\phi_1(x_1)} y_2,\\ 
        &\frac{d x_2}{dt} = \frac{1}{\phi_2(x_2)} \left( y_3 - y_4\right),\\ 
        &\frac{d x_3}{dt} = \frac{1}{\phi_3(x_3)} \left( y_4 - y_0\right),\\
        &x_0(t) = x_1(t) , \quad y_0(t) = y_1(t) + y_2(t), \quad y_4(t) = \frac{1}{10} \sqrt{x_2(t)} \label{eq:tank_known_alg}\\
        &y_0(t) = \frac{1}{5} x_0 x_3, \quad y_3(t) = \frac{1}{10} \sqrt{x_0(t)}, \label{eq:tank_unknown_terms}\\
        &y_0(t), y_1(t), y_3(t), y_4(t) \ge 0,
    \end{align}
\end{subequations}
which is an index-2 DAE. 
Here, $\phi_i(x_i)$ describes the area-height profile of the tanks, which is defined by their shape. We set $\phi_0(x_0) {=} 1/10, \phi_1(x_1) {=} 1/2, \phi_2(x_2) {=} 2,\phi_3(x_3) {=} 10$, i. e., all tanks have a constant area profiles, as is the case for cylinders, for example. However, the area of the different tanks varies. To test the methods proposed in this work, we assume no knowledge of \eqref{eq:tank_unknown_terms} and define the unknown terms as ${\bof{z}(t) {=} (y_0(t), y_3(t))^{\top}}$. We aim to learn a neural mapping ${\bof{f}_{NN}: \bof{x}(t) \mapsto \bof{z}(t)}$. For this, we have access to noisy observations of the differential states from a given number of trajectories with different initial conditions.

\subsubsection{Ablation study for `fully' simultaneous approach} \label{sec:4tank_ablation}

We investigate the importance of our proposed initialization and Hessian approximation on a small-scale training problem, where we have access to observations from three trajectories. We train our neural DAE with a neural network with two hidden layers with 30 neurons each, using the $\tanh$ activation function. There are three steps to the fully simultaneous method: trajectory initialization using \eqref{eq:smooth_init}, weight initialization \eqref{eq:nn_init}, and solution of \eqref{eq:colloc_NODE}, either using L-BFGS or exact Hessian information. We are interested in the overall training time when conducting only a subset of these steps, as well as the solution that the different configurations converge to. We define five different configurations (Trials A-E, cf. Table \ref{tab:ablation_tank}). We repeated the analysis on three different training problems, where different random noise on the observations and random initialization of the NN weights (before \eqref{eq:nn_init} was applied, if applicable) was used, and we  report the average.

\begin{figure}
        \centering
        \begin{subfigure}[b]{0.99\textwidth}
         \includegraphics[width=\textwidth]{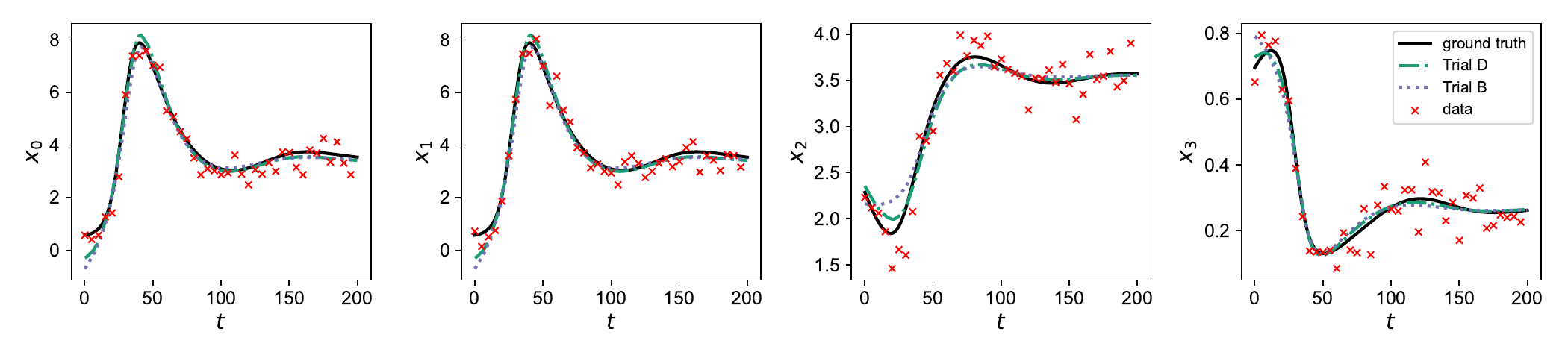}
         \caption{}
         \label{fig:ablation_traj_x}
     \end{subfigure}\\
     \begin{subfigure}[b]{0.5\textwidth}
         \includegraphics[width=\textwidth]{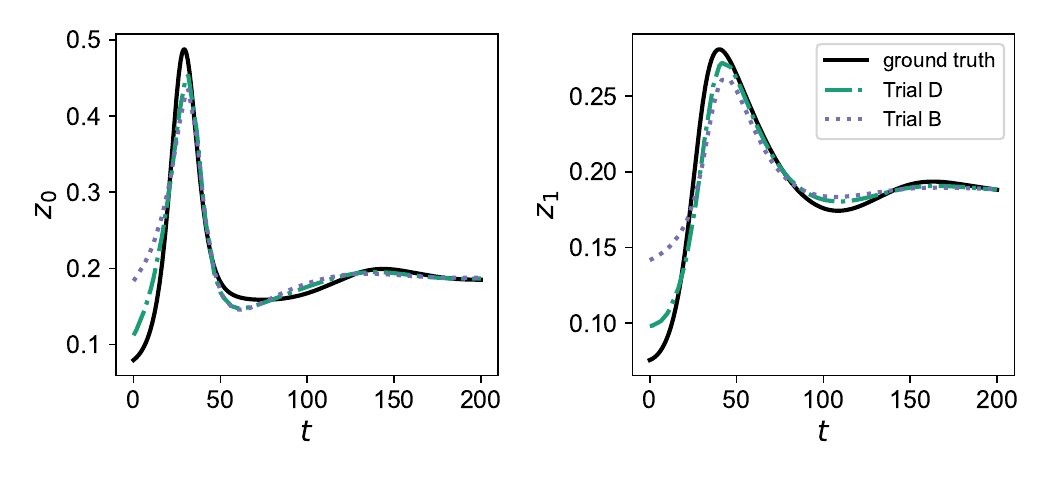}
         \caption{}
         \label{fig:ablation_traj_z}
     \end{subfigure}
     \caption{Comparison of trained hybrid models for different configurations from Table \ref{tab:ablation_tank}. Observed state trajectories (\ref{fig:ablation_traj_x}) and trajectories of the unknown terms (\ref{fig:ablation_traj_z}).}
    \label{fig:ablation_traj}
\end{figure}

For each step, we report wall-clock time. The mean-squared error (MSE) reported for training refers to the difference between the trajectories of the learned terms compared to ground truth -- this gives the most direct measure of the accuracy of the learned model. From Table \ref{tab:ablation_tank}, it appears that for the problem under consideration, not using any initialization scheme and using exact Hessian lead to the most accurate model, however at an unacceptable computational cost (trial C). Simply switching to L-BFGS approximation reduces the solution time substantially, although this proves to be less robust, as one problem could not be solved (trial D). Using the trajectory and/or weight initialization schemes resulted in less accurate models (trials A, B, E). For the problem considered here, the bias towards smooth output trajectory proves to be significant. This is confirmed in Figure \ref{fig:ablation_traj}, where we show the fit of the models trained with configurations D (no initialization) and B (weight and trajectory initialization) on a particular trajectory in the training set. Nevertheless, the initialization schemes are effective in reducing the number of required IPM iterations, as well as the robustness of the algorithm. The weight initialization only has minor effects on the result of the training procedure.

Note that there are several hyperparameters of our approach (weight regularization, smoothing penalty coefficient, discretization scheme), which affect the specific result from the study conducted here. The authors believe the results shown here to be representative for the problem at hand.

 \begin{table}
    \caption{Comparison of different configurations for the fully simultaneous approach }\label{tab:ablation_tank}
    \vspace{0.1cm}
    \begin{tabular}{c|ccccc|ccc}
    \toprule
     & \multicolumn{5}{c|}{\textbf{Step}} & \multicolumn{2}{c}{\textbf{Result}} \\
    \midrule
    \textbf{Trial} & \thead{Trajectory\\init. \eqref{eq:smooth_init}}
    & \thead{Weight\\init. \eqref{eq:nn_init}}
    & \thead{NLP \eqref{eq:colloc_NODE},\\L-BFGS}
    & \thead{NLP \eqref{eq:colloc_NODE},\\Exact}
    & Total
    & \thead{MSE\\ (train)}
    & \#Solved \\
    & (s) & (s) & (s)/\#iter. & (s)/\#iter. & (s) & \\
    \midrule
    A & \num{0.26} & \num{11.3} & - & \num{202.4}/51 & \num{213.9} & \num{0.087} & 3/3 \\
    B & \num{0.21} & \num{10.8} & \num{25.9}/108 & - &  \num{36.91} & \num{0.084} & 3/3 \\
    C & - & - & - & \num{711.0}/180 & \num{711.0} & \num{0.067} & 3/3\\
    D & - & - & \num{32.2}/134 & - & \num{32.2} & \num{0.072} & 2/3\\
    E & \num{0.22} & - & \num{27.7}/119 & - & \num{27.92} & \num{0.086} & 3/3\\ 
    \midrule
    \end{tabular}
\end{table}

\subsubsection{Scalability of decomposition approach}\label{sec:scalability_decomp}

In order to test whether the bi-level decomposition approach (cf. Section \ref{sec:decomp}) is viable, two aspects are of importance: the scalability in terms of the problem size and the ability to converge to a `good' solution. As opposed to the fully simultaneous approach, where termination criteria within \textsc{IPOPT} are well defined, there is no obvious stopping point for gradient descent methods, which is used to find an approximate solution to \eqref{eq:decomp_UP}. In this work, we used heuristic stopping criteria which are common in the machine learning community. Namely, we systematically lower the step length, or learning rate, of the gradient descent method and terminate once no further improvement in the objective is observed, or some maximum number of steps is reached. Furthermore, we use the popular Adam optimizer \citep{kingma2014adam} instead of vanilla gradient descent. All results shown in this section are averaged over three executions with varying realization of noise in the observed data and weight initialization.

\begin{figure}
\centering
\begin{adjustbox}{minipage=\linewidth,scale=1}
     \centering
          \begin{subfigure}[b]{0.45\textwidth}
         \centering
         \includegraphics[width=\textwidth]{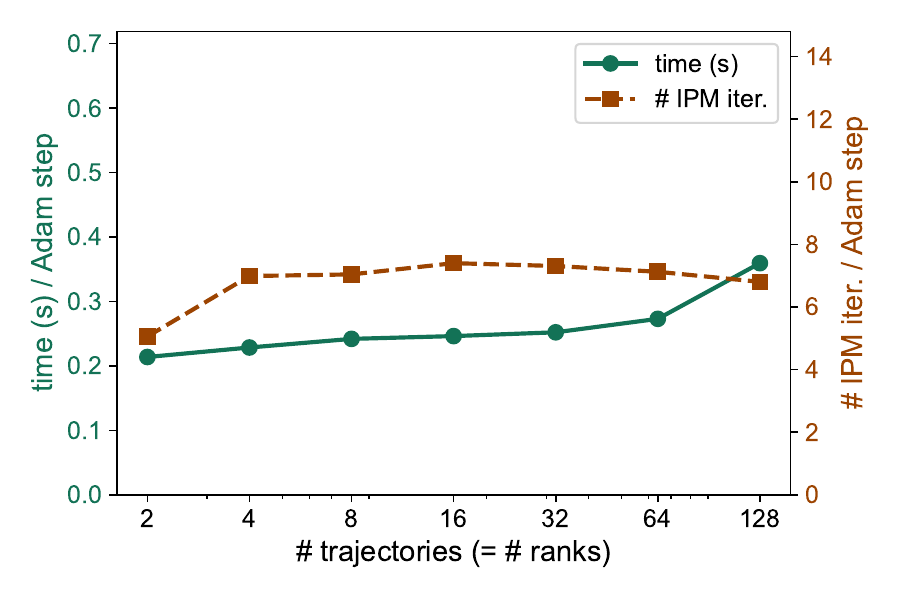}
         \caption{}
         \label{fig:decomp_scaling_nt_time}
     \end{subfigure}
     \hspace{0.6cm}
     \begin{subfigure}[b]{0.45\textwidth}
         \centering
         \includegraphics[width=\textwidth]{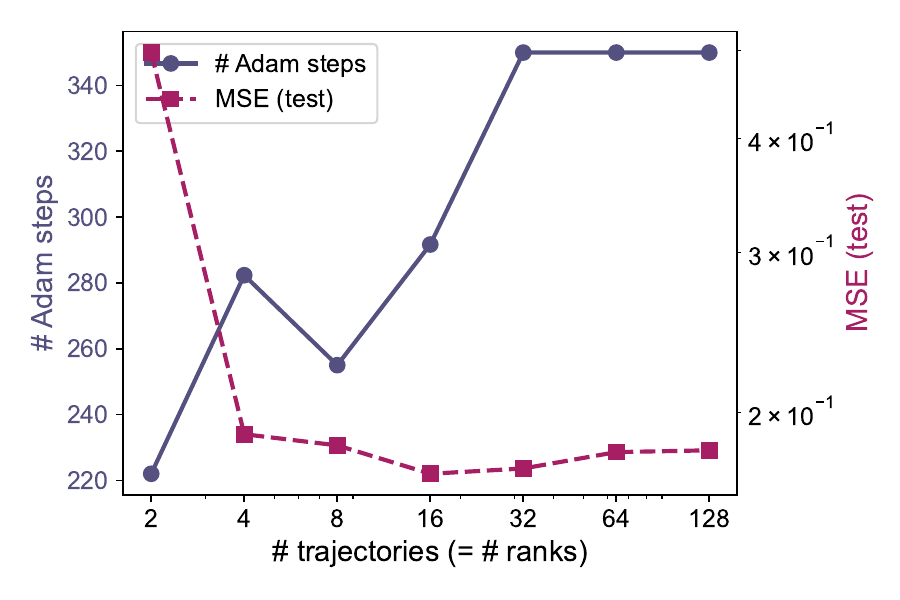}
         \caption{}
         \label{fig:decomp_scaling_nt_acc}
     \end{subfigure}
     \end{adjustbox}
     \caption{Scalability of bi-level decomposition with increasing amount of training data.}
    \label{fig:decomp_scaling_nt}
\end{figure}

We begin by testing the parallel scaling of the decomposition approach with respect to the number of trajectories (cf. Figure \ref{fig:decomp_scaling_nt}), where the size of the neural network is kept constant at three hidden layers with 50 neurons each. In Figure \ref{fig:decomp_scaling_nt_time}, we show that the time required per subproblem solution, as well as the associated number of IPM steps, does not change significantly as we increase the number of trajectories and ranks. This is expected, as the size of the subproblems does not change. When using all available processors on the node (128), there is a slight increase in solution time, most likely due to memory limitations. In Figure \ref{fig:decomp_scaling_nt_acc}, we show the number of Adam steps taken towards the solution of the upper problem \eqref{eq:decomp_UP}, as well as the test set accuracy of the of the learned component compared to ground truth. Again, these values are tracked as the number of trajectories is increased. Evidently, for fewer trajectories the gradient descent reaches its stopping criteria before reaching the pre-defined maximum number of steps (350). The resulting models have relatively poor generalization accuracy, which is expected for small training datasets. As the size of the problem is increased, the gradient descent algorithm requires more steps and eventually reaches the upper limit. It appears that algorithmically, the gradient descent method is likely to require more iterations as the amount of data is increased, which clearly impacts the overall runtime. Tuning the learning rate and stopping criteria could improve the performance on larger data sets.

\begin{figure}
\centering
\begin{adjustbox}{minipage=\linewidth,scale=1}
     \centering
          \begin{subfigure}[b]{0.45\textwidth}
         \centering
         \includegraphics[width=\textwidth]{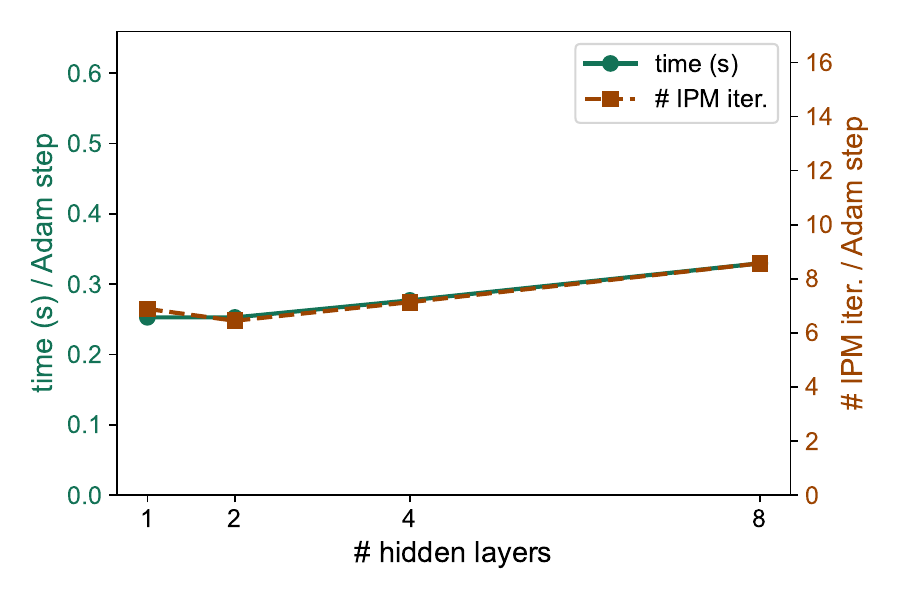}
         \caption{}
         \label{fig:decomp_scaling_nl_time}
     \end{subfigure}
     \hspace{0.6cm}
     \begin{subfigure}[b]{0.45\textwidth}
         \centering
         \includegraphics[width=\textwidth]{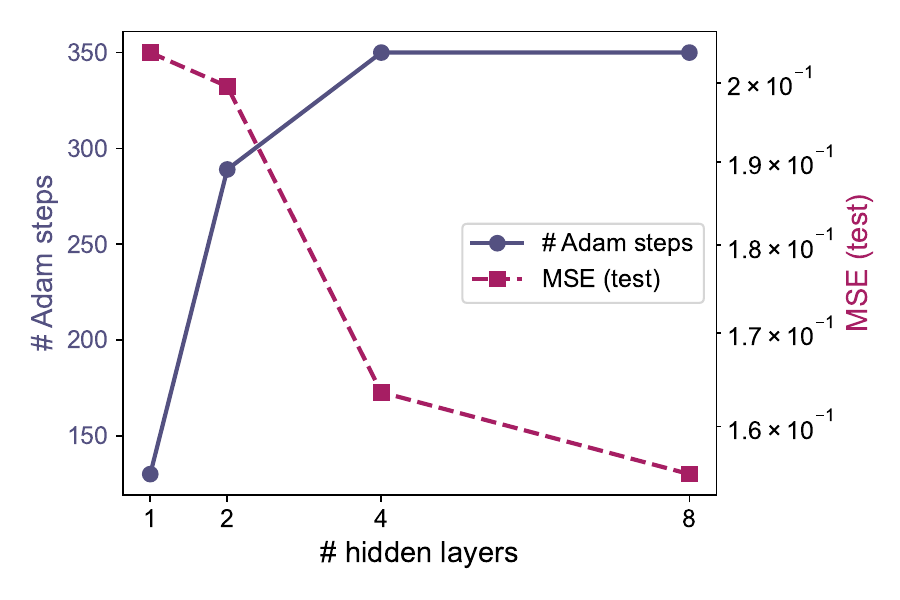}
         \caption{}
         \label{fig:decomp_scaling_nl_acc}
     \end{subfigure}
     \end{adjustbox}
     \caption{Scalability of bi-level decomposition with increasing size of the neural network.}
    \label{fig:decomp_scaling_nl}
\end{figure}

In addition, we investigate how the size of the neural network affects the performance of the decomposition scheme. For this we defined networks with 50 neurons per hidden layer and subsequently increase the number of layers. As before, we track elapsed time and IPM iterations per gradient step, total number of gradient descent steps and the accuracy of the trained models. The size of the data set is kept constant at 50 trajectories, while using 50 parallel cores. The results are shown in Figure \ref{fig:decomp_scaling_nl}. It appears that the time required per subproblem solution does not change significantly with the size of the neural network (cf. Figure \ref{fig:decomp_scaling_nl_time}). This is one of the intended advantages of the decomposition, as the neural network parameters are fixed for each subproblem solution. As before, we see that the number of Adam steps increases with increasing problem complexity (cf. Figure \ref{fig:decomp_scaling_nl_acc}). Furthermore, larger neural networks result in improved test accuracy of the learned models.

\subsubsection{Scalability of fully simultaneous approach}\label{sec:scalability_simul}

We perform similar scalability tests for the fully simultaneous approach. It has to be noted that this method was not able to scale to the problem sizes considered in Section \ref{sec:scalability_decomp}, as it is not parallelized and memory limitations were observed when attempting to apply the method to the training of larger neural networks, even when employing the gray-box formulation for the neural network components. Hence we repeat the analysis from Section \ref{sec:scalability_decomp} on a smaller scale. We solely focus on the solution time of the NLP \eqref{eq:colloc_NODE}, as the initialization steps are negligible in terms of the overall training time. We use the L-BFGS approximation for all tests in this section.

\begin{figure}
\centering
\begin{adjustbox}{minipage=\linewidth,scale=1}
     \centering
          \begin{subfigure}[b]{0.45\textwidth}
         \centering
         \includegraphics[width=\textwidth]{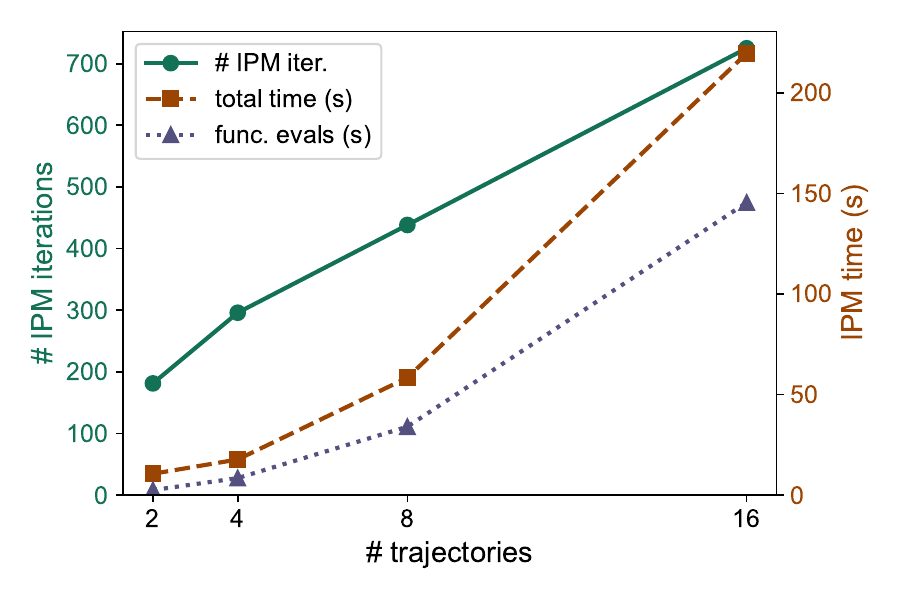}
         \caption{}
         \label{fig:simul_scaling_nt_time}
     \end{subfigure}
     \hspace{0.6cm}
    \begin{subfigure}[b]{0.45\textwidth}
         \centering
         \includegraphics[width=\textwidth]{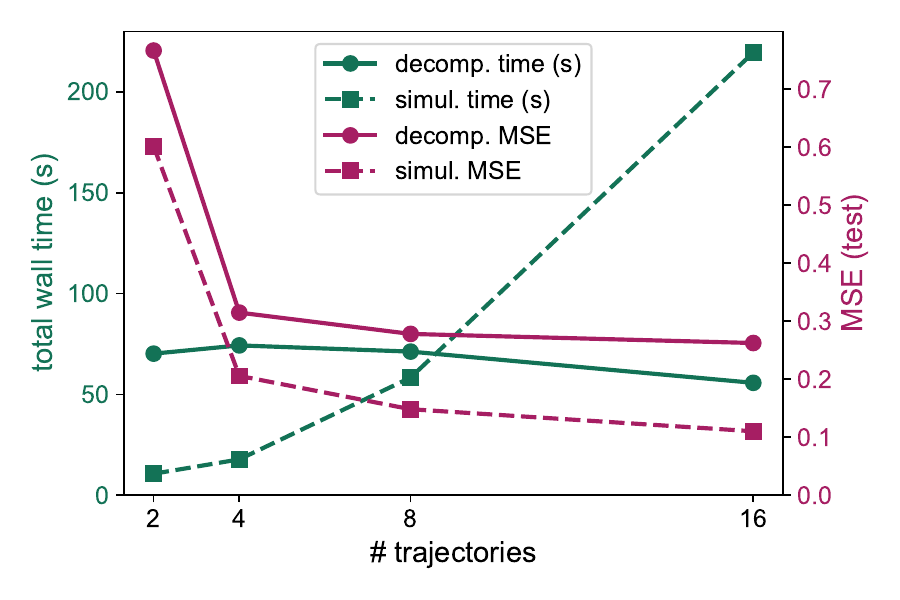}
         \caption{}
         \label{fig:simul_decomp_comparison}
     \end{subfigure}
     
     \begin{subfigure}[b]{0.45\textwidth}
         \centering
         \includegraphics[width=\textwidth]{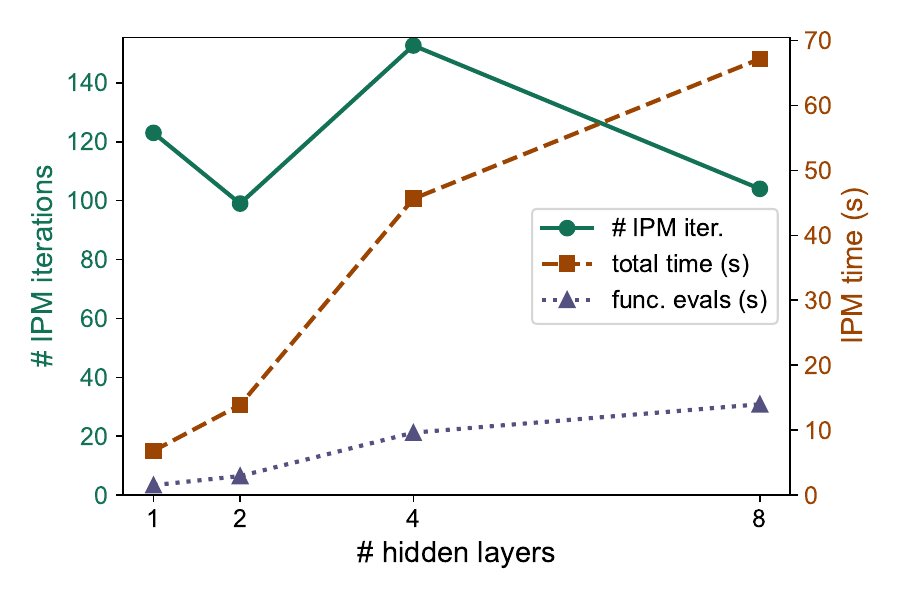}
         \caption{}
         \label{fig:simul_scaling_nl_time}
     \end{subfigure}

\end{adjustbox}
     \caption{Scalability of the fully simultaneous approach when increasing the number of trajectories (a). Comparison of total runtime and model accuracy for the simultaneous and decomposition approach (b). Scalability of the fully simultaneous approach when increasing the size of the trained model (c).
     }
    \label{fig:simul_scaling}
\end{figure}

To test the scalability with respect to the number of trajectories, we used a neural network with two layers and ten neurons each. The training problem was then solved while increasing the number of included trajectories from two to 16. The time required to solve the training problem using \textsc{IPOPT}, as well as the number of iterations are shown in Figure \ref{fig:simul_scaling_nt_time}. As expected, the overall solution time increases significantly, as does the number of iterations. A significant portion of the solution time is spent evaluating functions and derivatives required for the KKT system at every step, which includes the evaluation of the neural network component. It appears that the vectorization across evaluation points using the gray-box model does not result in the expected insensitivity in evaluation time with respect to the number of those points.

For the same experimental setup, we compare the total time required for training and the test set accuracy of the resulting models between the fully simultaneous and decomposition approaches (cf. Figure \ref{fig:simul_decomp_comparison}). As expected, the solution time of the simultaneous approach quickly surpasses that required for the decomposition, as the number of trajectories increases. However, the accuracy of the models trained using the simultaneous approach is significantly higher, even when compared to the larger models produced by the decomposition approach on larger training sets in Section \ref{sec:scalability_decomp}. This points towards the fact that (a) the gradient descent configuration for the decomposition might be significantly tuned (b) for the problem considered here, there might not be much value in more training data, as also indicated by Figure \ref{fig:decomp_scaling_nt_acc}. Nevertheless, the ability of the fully simultaneous approach to find high-quality local solutions for small training problems is worth noting.

Lastly, we evaluate scalability of the fully simultaneous approach when increasing the number of layers in the trained model (cf. Figure \ref{fig:simul_scaling_nl_time}), where the number of neurons per layer is fixed at ten. Here, we only consider two trajectories for each problem. Interestingly, the number of iterations required by the solver does not increase significantly with more layers, although the overall solution time still does. This means that the time per iteration increases significantly. In combination with the relatively low portion of time spent in function evaluations, this points towards the fact that the factorization of the linear system at every step becomes more costly as the size of the neural network is increased. This is plausible as the sparsity of the system decreases for larger neural networks, which negatively impacts the performance of the linear solver.

\subsection{Population dynamics}\label{sec:pop_dyn}

The following ODE describes the dynamics of a general predator-prey system, and is frequently used as a baseline to analyze population dynamics \citep{bazykin1998nonlinear}:
\begin{subequations}\label{eq:ode_pop}
    \begin{align}
        &\frac{d x_0}{dt} = (r_1 - a_1  x_1 - b_1 x_0)x_0, \\
        &\frac{d x_1}{dt} = (r_2 - a_2 \frac{x_1}{x_0})x_1, \label{eq:ode_pop2}
    \end{align}
\end{subequations}
where $x_0(t)$ and $x_1(t)$ describe the population of the prey and predator, respectively. The constant, positive parameters $a_1{=}1/5, a_2=1/100, r_1{=}1/5, r_2{=}1/5, b_1{=}1/10$ describe the behavior of the populations. As a test case for our hybrid model, we assume that the term $z(t){=}r_2 - a_2 \frac{x_1}{x_0}$ in \eqref{eq:ode_pop2} is unknown and apply the approach described in Sec. \ref{sec:proposed_approach} to find a mapping ${\bof{f}_{NN}: \bof{x}(t) \mapsto z(t)}$.
Furthermore, we assume knowledge of the following Lyapunov function for \eqref{eq:ode_pop} \citep{korobeinikov2001lyapunov}:
\begin{equation}
    V = \ln \left(\frac{x_0}{x_0^*}\right) + \frac{x_0^*}{x_0} + \frac{a_1 x_0^*}{a_2}\left(\ln\left(\frac{x_1}{x_1^*}\right) + \frac{x_1^*}{x_1}\right),
\end{equation}
where $x_0^* {=} \frac{r_1a_2}{a_1r_2 + a_2b_1}$ and  $x_1^* {=} \frac{r_1r_2}{a_1r_2 + a_2b_1}$ denote the fixed point of the system. We use this to define $V(t)$ as a differential variable in the training problem and impose a path constraint $\frac{dV}{dt} \le 0$. This enforces the state trajectories learned by our method to adhere to the Lyapunov function, which is expected to yield learned mappings that generalize better.

\subsubsection{Effect of smoothing penalty}

\begin{figure}
\centering
\begin{adjustbox}{minipage=\linewidth,scale=1}
     \centering
          \begin{subfigure}[b]{0.31\textwidth}
         \centering
         \includegraphics[width=\textwidth]{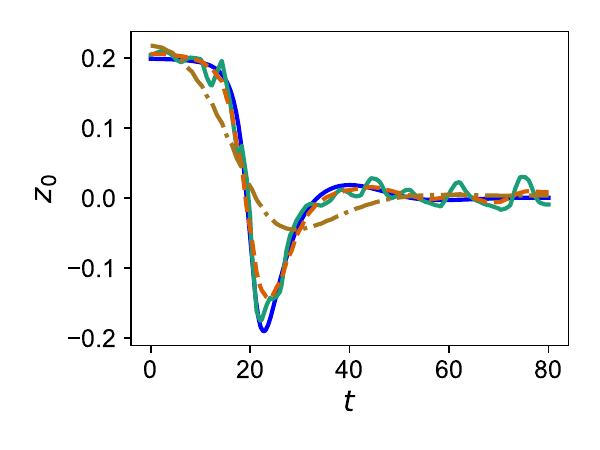}
         \caption{}
         \label{fig:pop_z_smooth}
     \end{subfigure}
     \hspace{0.6cm}
     \begin{subfigure}[b]{0.62\textwidth}
         \centering
         \includegraphics[width=\textwidth]{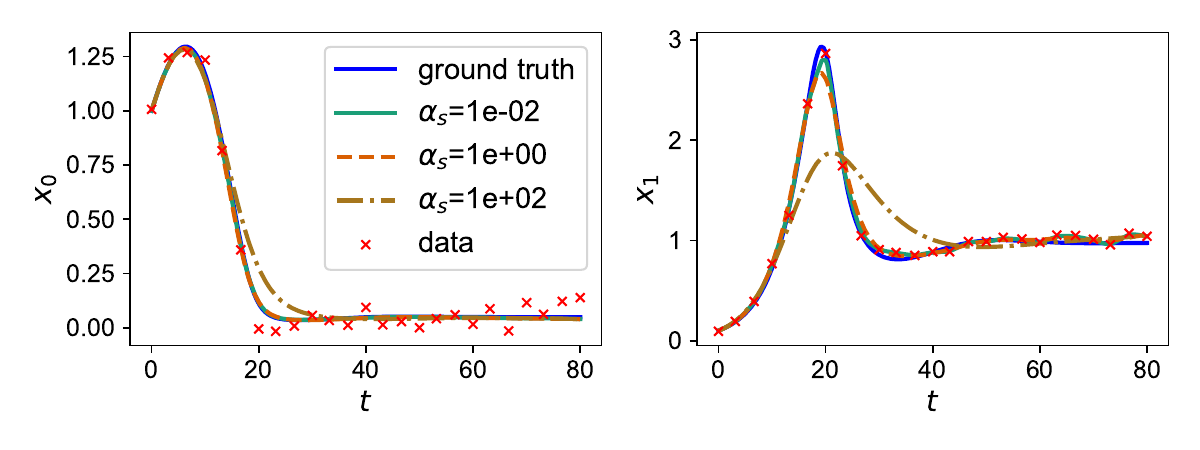}
         \caption{}
         \label{fig:pop_states_smooth}
     \end{subfigure}
     \end{adjustbox}
     \caption{Smoothed trajectory of unknown term $z(t)$ (a), and states $x(t)$ (b), for varying smoothing penalties $\alpha_s$. The observations are shown in red.}
    \label{fig:pop_smooth}
\end{figure}

In Fig. \ref{fig:pop_smooth}, we illustrate how the solutions of the smooth initialization problem \eqref{eq:smooth_init} are affected by the choice of smoothing coefficient $\alpha_s$. For a single observed trajectory, we plot the resulting trajectories for the states and unknown terms. Note that the neural map for the unknown terms has not been included into the problem formulation at this point. It is evident that a low smoothing coefficient results in jagged trajectories, whereas increasing it produces smooth trajectories which fit the observations less accurately.

\subsubsection{Effect of Lyapunov-based path constraints}

\begin{figure}
\centering
\begin{adjustbox}{minipage=\linewidth,scale=0.75}
        \begin{subfigure}[b]{0.27\textwidth}
         \centering
         \includegraphics[width=\textwidth]{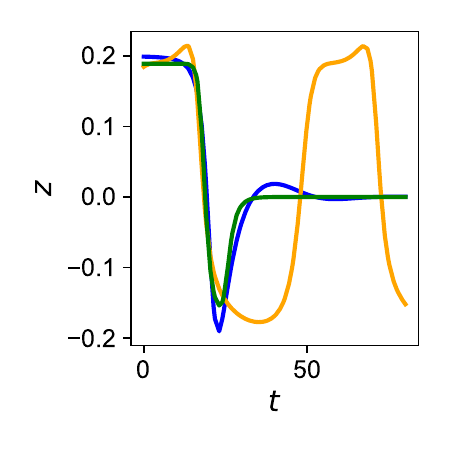}
         \caption{}
         \label{fig:pop_z_lyap}
     \end{subfigure}
     \begin{subfigure}[b]{0.63\textwidth}
         \centering
         \includegraphics[width=\textwidth]{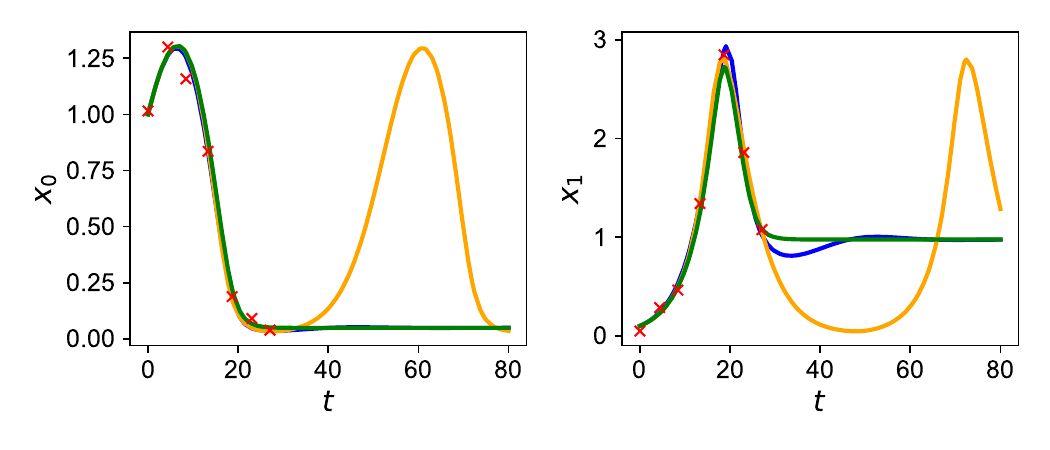}
         \caption{}
         \label{fig:pop_states_lyap}
     \end{subfigure}\\
     \begin{subfigure}[b]{0.3\textwidth}
         \centering
         \includegraphics[width=\textwidth]{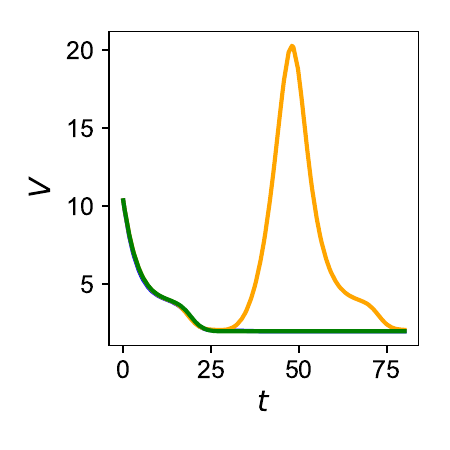}
         \caption{}
         \label{fig:pop_V_lyap}
     \end{subfigure}
     \begin{subfigure}[b]{0.6\textwidth}
         \centering
         \includegraphics[width=\textwidth]{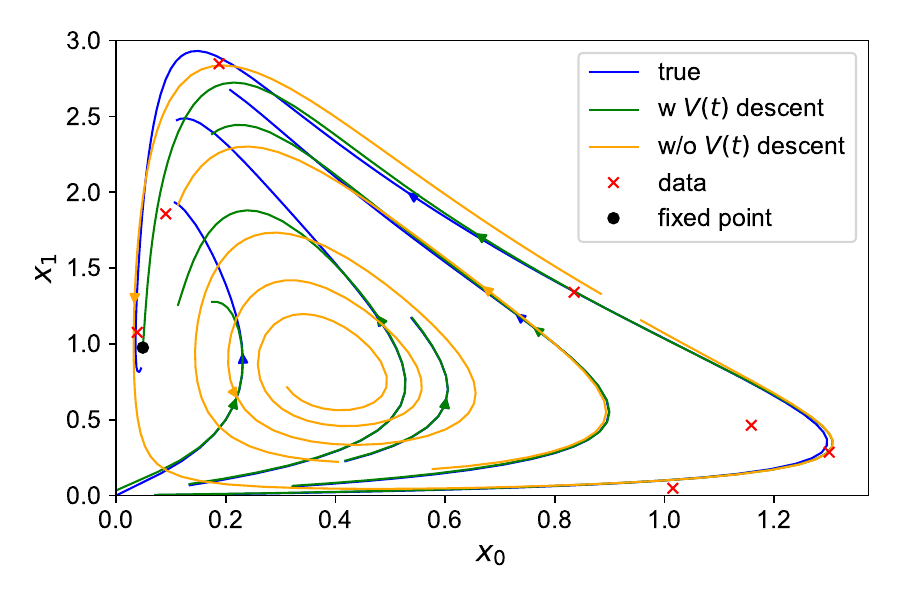}
         \caption{}
         \label{fig:pop_lyap_stream}
     \end{subfigure}
     \end{adjustbox}
     \caption{State trajectory of the hybrid systems, which are learned with and without the path constraint enforcing Lyapunov descent (\ref{fig:pop_states_lyap}). Including the constraints leads to a more accurate model, especially beyond the point where data are observed. The output of the neural map (\ref{fig:pop_z_lyap}) shows similar improvement when compared with the true model.}
    \label{fig:pop_lyap}
    
\end{figure}

We now demonstrate how including physically-motivated path constraints, such as the enforcing the descent of the Lyapunov function in \eqref{eq:ode_pop}, can significantly improve a learned hybrid model, especially in cases where little data are available. In Fig. \ref{fig:pop_states_lyap}, the trajectories of two models fitted with our approach are plotted, one with and one without the path constraint. Including the constraint leads to a model which has much improved generalization capabilities, as shown by the phase plot produced by the two learned models in Fig. \ref{fig:pop_lyap_stream}. Without including the constraint, the descent property of the learned model is clearly violated (\ref{fig:pop_V_lyap}). This example is perhaps slightly contrived, as knowing the Lyapunov function is unlikely in cases where the system dynamics themselves are not fully known. However, this case study succinctly demonstrates the flexibility of the simultaneous approach in dealing with non-trivial constraints. In this example, we used a neural network with two hidden layers, each with 10 neurons using the $\tanh$ activation function.

\subsection{Fed-batch bioreactor}\label{sec:fedbatch}
We consider a model of a fed-batch bioreactor, adapted from \citet{cbe30338_github}. In this system, a generic cell bioproduct is generated, which is a relevant case study in the area of bio-pharmaceutics. The ODEs for cell concentration $X \, [g/L]$, desired product concentration $P \, [g/L]$, necessary substrate concentration $S \, [g/L]$, and volume $V \, [L/h]$ are given by:
\begin{subequations}\label{eq:fedbatch}
\begin{align}
    &\frac{d X}{d t} =-\frac{F(t)}{V(t)} \cdot X(t) + \mu(S(t)) \cdot X(t) \label{eq:fedbatch01} \\ 
    &\frac{d P}{d t} =-\frac{F(t)}{V(t)} \cdot P(t) + Y_{P / X} \cdot\mu(S(t))\cdot X(t) \\
    &\frac{d S}{d t} =\frac{F(t)}{V(t)}\cdot\left(S_f-S(t)\right)-\frac{1}{Y_{X / S}}\cdot \mu(S(t)) \cdot X(t) \label{eq:fedbatch03}\\ 
    &\frac{d V}{d t} =F(t) \\[1mm]
    &X(t), P(t), S(t), V(t) \ge 0, \quad \forall_{t\in [t_0, t_f]}\label{eq:fedbatch_lb}
\end{align}
\end{subequations}
where $F(t) \, [L/h]$ corresponds to a feed profile that maintains the substrate at a viable concentration for the cell growth. 
For a Monod-type kinetic model, we have
\begin{equation} \label{eq:Monod}
\mu(S(t))=\mu_{\max } \frac{S(t)}{K_S+S(t)},
\end{equation}
where $\mu_{max} [h^{-1}] $ is maximum specific growth rate and $K_s [\frac{g}{L}]$ the half-saturation kinetic constant. Other parameters of the system include the feed substrate concentration ${S_f {=} S(t_0)}$, the yield coefficient for new cells $Y_{X/S}$ and the product yield coefficient, $Y_{P/X}$. We adopt these parameters (except the initial state) from \citep{cbe30338_github}.

In this example, we seek to learn a relationship ${f_{\text{NN}}: (X, P, S, V) \mapsto \mu}$ from observations of the state variables, i. e., we assume no knowledge of \eqref{eq:Monod}. For this example, we wish to focus on the inference task. Once $f_{\text{NN}}$ is trained, it can be embedded into \eqref{eq:fedbatch}, and the resulting ODE can be evaluated on new initial conditions. However, when using an integrator directly, the resulting trajectories are not guaranteed to obey physical limits, such as the lower bounds on the state variables \eqref{eq:fedbatch_lb}. This applies regardless of which method was used to train $f_{\text{NN}}$, although there might be differences in the frequency at which such bounds are violated. Alternatively, the simultaneous neural DAE framework described in this work can also be used for inference tasks. For the system at hand, the following NLP is formulated and discretized using the approach outlined in Section \ref{sec:proposed_approach}:
\begin{subequations}\label{eq:fb_inference_nlp}
    \begin{align}
        \min &\int_{t_0}^{t_f} |\Delta^{+}(t) + \Delta^{-}(t)| dt \\
        \text{s. t.} \quad &\frac{d X}{d t} =-\frac{F(t)}{V(t)}\cdot X(t) + z(t) \cdot X(t)  \\ 
    &\frac{d P}{d t} =-\frac{F(t)}{V(t)} P(t) + Y_{P / X}\cdot z(t) \cdot X(t)\\
    &\frac{d S}{d t} =\frac{F(t)}{V(t)}\cdot\left(S_f-S(t)\right)-\frac{z(t)}{Y_{X / S}}\cdot X(t) \\ 
    &\frac{d V}{d t} =F(t) \\
    &z(t) = f_{\text{NN}}(X(t), P(t), S(t), V(t), \bs{\theta}) + \Delta^{+}(t) - \Delta^{-}(t) \\
    &X(t), P(t), S(t), V(t), \Delta^{+}(t), \Delta^{-}(t) \ge 0, \quad \forall_{t\in [t_0, t_f]}.
    \end{align}
\end{subequations}
Note that for inference, $\bs{\theta}$ is fixed. Furthermore, we have added slack variables to the constraint defining the learned relationship. This allows the solution of \eqref{eq:fb_inference_nlp} to obey other constraints defined in the system, e.g. lower bounds on the state variables by modulating the output of the trained neural network when necessary.

\begin{figure}
    \centering
        \begin{subfigure}[b]{0.99\textwidth}
         \includegraphics[width=\textwidth]{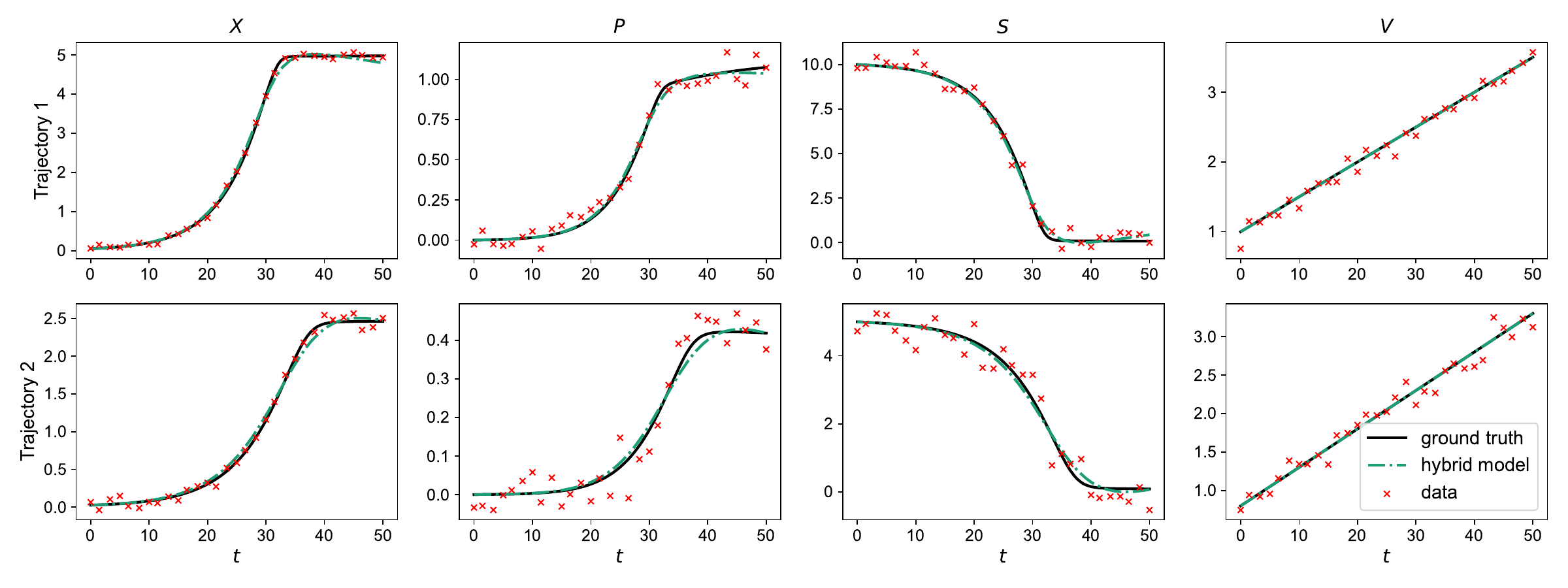}
         \caption{}
         \label{fig:fb_x_train}
     \end{subfigure}\\
       \begin{subfigure}[b]{0.99\textwidth}
         \includegraphics[width=\textwidth]{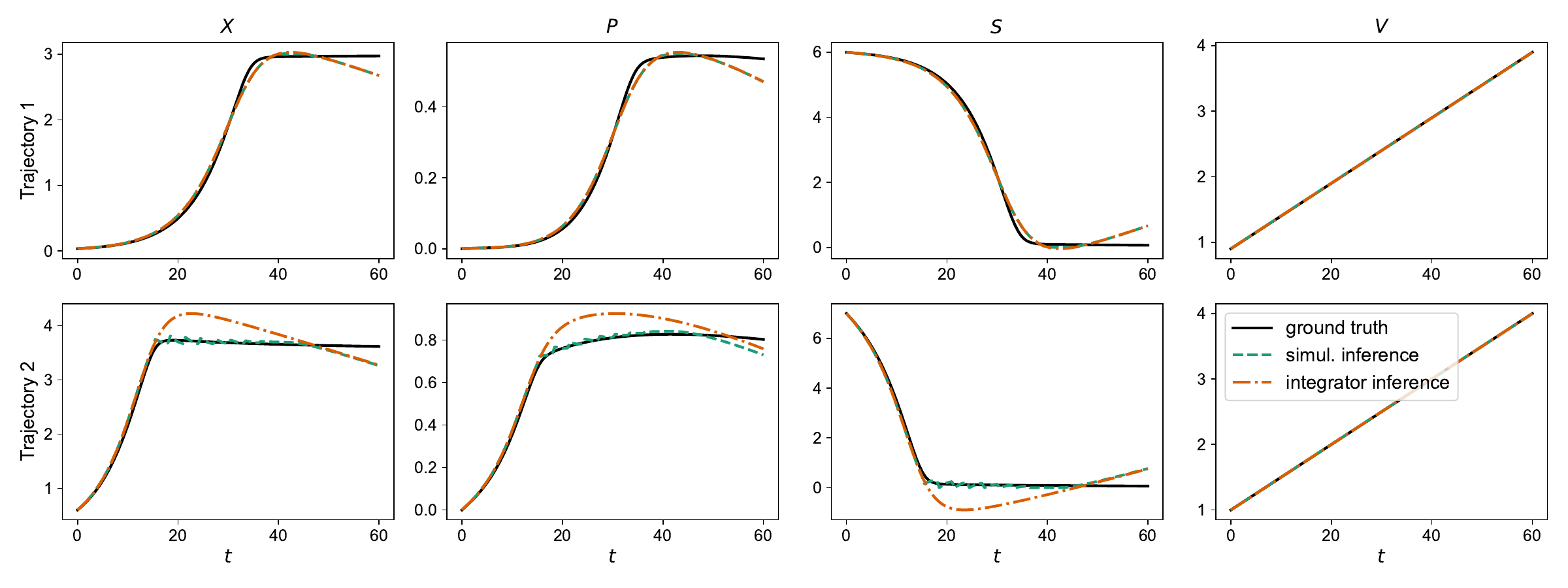}
         \caption{}
         \label{fig:fb_x_infer}
     \end{subfigure}\\
     \caption{A hybrid model is trained on two trajectories (a) and evaluated on two new initial conditions (b). Performing inference using the simultaneous approach ensures known constraints (here, $S(t) \ge 0 )$ are respected.}
\end{figure}

To illustrate this, a hybrid model was trained using the fully simultaneous approach in Section \ref{sec:proposed_approach}, although any of the other approaches can be applied as well. The model was trained on only two trajectories, where it achieved a good fit (cf. Figure \ref{fig:fb_x_train}). Due to few training data, the trained model is not likely to generalize well. In Figure \ref{fig:fb_x_infer}, we evaluate the trained hybrid ODE on two unseen initial conditions. We use both a standard integrator (Runge-Kutta 45 from \textsc{Scipy} \citep{virtanen2020scipy}) and the simultaneous approach \eqref{eq:fb_inference_nlp} to compute the trajectories for the learned model. The results show that the the trajectories produced by the two approaches agree, when none of the constraints would be violated (Trajectory 1 in Figure \ref{fig:fb_x_infer}). However, the simultaneous inference approach ensures the satisfaction of bound constraints, even where the straightforward integration of the learned model does not (Trajectory 2 in Figure \ref{fig:fb_x_infer}), and thereby produces a trajectory closer to ground truth, in this example. For simple problems like the one considered here, integration schemes can be modified to achieve similar effects. However, for the larger class of DAEs, it is usually less straight-forward to obey algebraic constraints. The evaluation of the inference scheme outlined here on a wider range of problems appears to be an interesting area for further research.

\section{Conclusion} \label{sec:conclusion}

In this work, we investigate the applicability of the simultaneous approach for dynamic optimization to the training of so-called neural differential-algebraic systems of equations (neural DAEs). We adopt a modeling framework for neural DAEs which focuses on hybrid models, i. e. systems where only a subset of the differential or algebraic equations is approximated with a neural network, which is trained on observed data. We formulate the training task as a DAE-constrained optimization problem, and propose the use of the simultaneous approach to produce a discretized nonlinear programming problem (NLP), which is solved using an NLP solver, such as \textsc{IPOPT}. To improve the tractability of the resulting NLP, we devise a specialized initialization scheme, as well as the use of existing solver and modeling configurations, namely L-BFGS Hessian approximations and gray-box models, respectively.

Furthermore, we propose a bi-level decomposition approach to tackle the scalability issues of the simultaneous approach with respect to the size of the neural network and the amount of training data. On the lower level, this leverages implicit differentiation of the solutions of the subproblems with respect to the neural network parameters, which are updated on the upper level using a gradient descent method.

In our computational experiments, we demonstrate that the simultaneous approach can effectively solve training problems of small to moderate sizes, including on higher-index DAE systems. In a scalability study, it is shown that the decomposition approach is able to scale to much larger problems through the parallel evaluation of subproblems and the separation of the neural network parameters from the solution of the NLPs. As is typical with first-order gradient-based training methods, hyperparameters and stopping criteria need to be chosen, which often requires problem-specific tuning. In fact, the fully simultaneous approach is shown to converge to better solutions than the sequential bi-level decomposition on small training problems.

Finally, there are several important avenues for future research and improvements. The efficacy of the methods proposed here should be evaluated on a wider range of systems, potentially including data from real experimental testbeds to assess the value of DAE-based hybrid models. The applicability of the simultaneous approach to inference tasks involving neural DAEs could also be studied in more detail. Furthermore, the implementation of the decomposition approach could be improved by integrating the NLP solver and implicit differentiation more closely, e. g. by making use of existing tools to evaluate sensitivity in \textsc{IPOPT} \citep{biegler2018large, pirnay2012optimal}. Furthermore, automatic differentiation and NLP solution could be conducted on graphical processing units to scale to larger learning tasks -- this has been an active area of development in recent years  \citep{parker2025nonlinear}.

\clearpage
\bibliography{sn-bibliography}

\end{document}